\documentclass{article} 
\usepackage{colm2024_conference}
\usepackage{booktabs}
\usepackage{graphicx}
\usepackage{enumitem}
\usepackage{wrapfig}
\usepackage{algorithm}
\usepackage{algpseudocode}
\usepackage{natbib}
\usepackage{makecell}
\usepackage{booktabs}
\usepackage{array}
\usepackage{amsmath}
\usepackage{amssymb}
\usepackage{amsfonts}
\usepackage{multirow}
\usepackage{verbatim}
\usepackage{caption}
\usepackage{longtable}
\usepackage{supertabular}
\usepackage{CJKutf8}
\usepackage[utf8]{inputenc} 
\usepackage[T1]{fontenc}
\usepackage[french,vietnamese,mongolian,greek,english]{babel}
\usepackage{pifont}
\usepackage{tabularx}
\usepackage{geometry}
\usepackage{enumitem}
\usepackage{tablefootnote}
\usepackage{xspace}
\usepackage{textcomp}
\usepackage{makecell}
\usepackage{lscape} 
\usepackage{siunitx}
\usepackage{listings}
\usepackage{xcolor}
\usepackage{svg}
\definecolor{dt}{gray}{0.7}
\usepackage{pifont}       
\usepackage{bbding}       
\usepackage{fontawesome}

\usepackage{scrextend}

\usepackage{tgpagella}
\usepackage{latexsym}
\usepackage[T1]{fontenc}
\usepackage[utf8]{inputenc}
\usepackage{microtype}
\definecolor{mydarkblue}{rgb}{0,0.08,0.45}
\definecolor{citecolor}{HTML}{0071BC}
\usepackage{url}            
\usepackage{nicefrac}       
\usepackage{changepage}
\usepackage{xargs}          
\usepackage{wrapfig,lipsum,booktabs}
\usepackage{longtable}
\usepackage{subcaption}
\usepackage{endnotes}

\usepackage{pgfplots}
\usetikzlibrary{pgfplots.groupplots}
\pgfplotsset{compat=1.3}
\usepackage{tikz}
\usetikzlibrary{patterns}

\usepackage[most]{tcolorbox}

\usepackage{hyperref}
\definecolor{darkblue}{rgb}{0, 0, 0.5}
\hypersetup{colorlinks=true, citecolor=darkblue, linkcolor=darkblue, urlcolor=darkblue, bookmarks=true, bookmarksopen=false, bookmarksnumbered=true}

\usepackage[capitalize,noabbrev]{cleveref}

\crefname{section}{\S}{\S\S}
\Crefname{section}{\S}{\S\S}
\crefname{subsection}{\S\S}{\S\S}
\Crefname{subsection}{\S\S}{\S\S}
\crefformat{section}{\S#2#1#3}
\crefformat{subsection}{\S\S#2#1#3}

\crefname{table}{Table}{Tables}
\crefname{figure}{Figure}{Figures}
\crefname{algorithm}{Algorithm}{}
\crefname{equation}{eq.}{}
\crefname{appendix}{Appendix}{}
\crefformat{section}{Section #2#1#3}
\usepackage{multicol}
\usepackage{tcolorbox}
\usepackage{titlesec}
\titleformat*{\section}{\large\bfseries}
\usepackage{array} 
\newcolumntype{P}[1]{>{\centering\arraybackslash}p{#1}} 
\usepackage{multirow}
\usepackage[table]{xcolor}

\usepackage{adjustbox}
\definecolor{objblue}{RGB}{3,139,221}  
\definecolor{attrred}{RGB}{255,67,67}    
\definecolor{easygreen}{RGB}{0,156,75}  
\definecolor{middleyellow}{RGB}{242,89,34}  
\definecolor{hardred}{RGB}{216,56,58}
\usepackage{colortbl}
\usepackage{array}

\usepackage[most]{tcolorbox}
\definecolor{BoxBackground}{RGB}{240, 240, 240} 
\definecolor{BoxFrame}{RGB}{0, 0, 0} 
\definecolor{TitleBackground}{RGB}{0, 0, 0} 
\definecolor{TitleText}{RGB}{255, 255, 255} 
\tcbset{
  academicbox/.style={
    boxsep=5pt,
    left=2pt,
    right=2pt,
    bottom=0.5pt,
    boxrule=0.5pt,
    colback=BoxBackground,
    colframe=BoxFrame,
    colbacktitle=TitleBackground,
    coltitle=TitleText,
    enhanced,
    attach boxed title to top left={yshift=-0.1in,xshift=0.1in},
    boxed title style={boxrule=0pt,colframe=white},
    title={#1},
  }
}
\newtcolorbox{AcademicBox}[1][]{academicbox=#1}

\title{Qwen-Image-Agent: Bridging the Context Gap \\in Real-World Image Generation}

\author{
Zekai Zhang, Jiahao Li, Jie Zhang, Kaiyuan Gao, Kun Yan, Lihan Jiang, \\ Ningyuan Tang, Shengming Yin, Tianhe Wu, Xiaoyue Chen, Xiao Xu, \\
Yan Shu, Yanran Zhang, Yixian Xu, Yuxiang Chen, Zhendong Wang, \\
Zihao Liu, Zikai Zhou, Huishuai Zhang, Dongyan Zhao, Chenfei Wu\thanks{Corresponding Author.}}

\begin{document}

\maketitle
\vspace{-3mm}

\begin{abstract}

While text-to-image (T2I) models have achieved remarkable progress, they struggle with real-world requests that are often underspecified, implicit, or dependent on up-to-date knowledge. We identify this challenge as the Context Gap: the mismatch between the user context and the sufficient generation context for T2I models. To bridge this gap, we propose Qwen-Image-Agent, a unified agentic framework that integrates plan, reason, search, memory and feedback in a context-centric manner. Qwen-Image-Agent treats user input as partial context and progressively constructs the generation context through Context-Aware Planning and Context Grounding. Specifically, Context-Aware Planning identifies missing context and plans how it should be acquired and used, while Context Grounding gathers this context from reason, search, memory, and feedback. To evaluate agentic image generation, we further introduce Image Agent Bench (IA-Bench), a benchmark covering four core image agent capabilities: Plan, Reason, Search, and Memory. Experiments on IA-Bench, Mindbench and WISE-Verified show that Qwen-Image-Agent outperforms strong baselines and achieves state-of-the-art performance.
\end{abstract}

\begin{figure}[h!]
    \centering
    \includegraphics[width=0.847\textwidth]{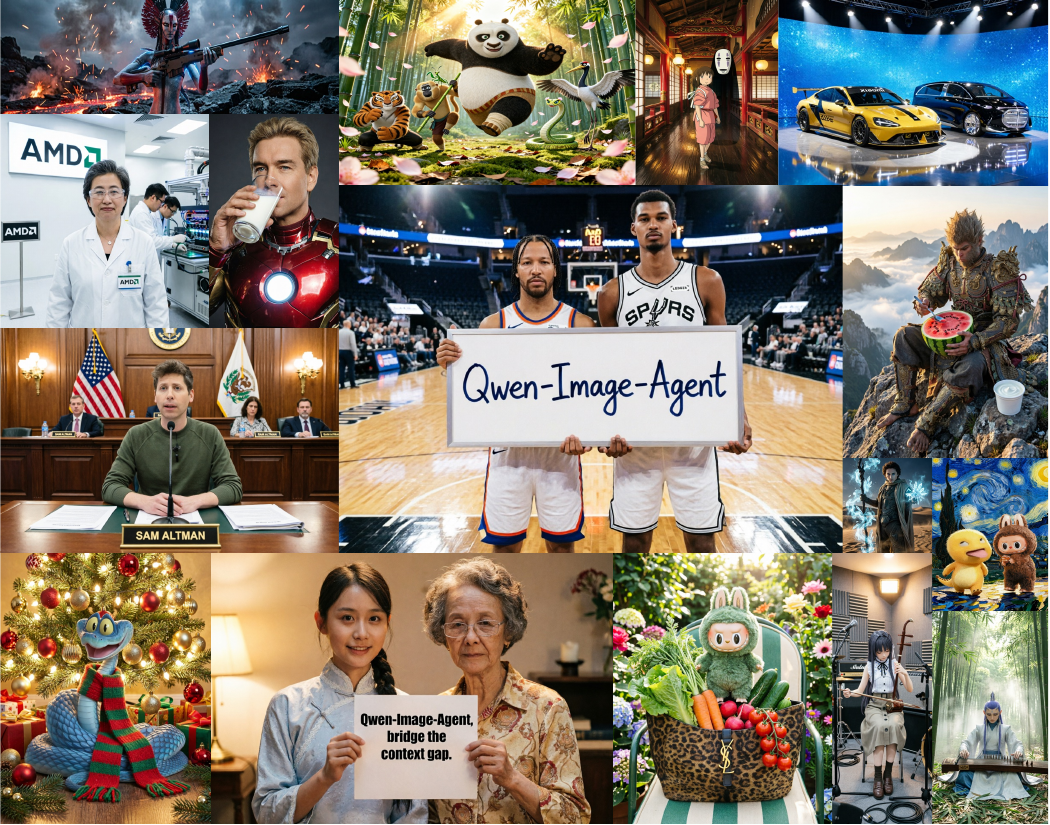}
    \caption{Qwen-Image-Agent examples, generated without providing visual references.}
    \label{fig:teaser}
\end{figure}

\newpage
\section{Introduction}

Text-to-image (T2I) models have achieved impressive progress in generating high-quality images from natural language prompts~\citep{flux2024, sd35Medium, Wu2025QwenImageTR}. As these systems move into real-world applications such as marketing, product design, and slide creation, they are increasingly expected to solve practical visual tasks rather than merely render prompts.

Despite their generative ability, current T2I models remain limited on real-world tasks~\citep{He2026MindBrushIA}. A key reason is a structural mismatch between training and deployment: models are optimized for fully specified prompts~\citep{Wu2025QwenImageTR}, while real-world requests are often underspecified. In practice, successful generation may require inferring implicit user intent, retrieving up-to-date knowledge or visual references from web, and incorporating interaction history.

We refer to this mismatch as the \textbf{Context Gap}: the gap between the provided \textbf{\textit{user context}} and the \textbf{\textit{generation context}} required for T2I models. This gap motivates a paradigm shift from traditional \textbf{\textit{direct image generation}} to \textbf{\textit{agentic image generation}}, where the system must identify missing context, acquire it, and use it effectively during generation. Recent work has explored components such as plan~\citep{Yao2026PhotoAgentAP}, reason~\citep{He2026MindBrushIA}, search and tool use~\citep{Ye2026AgentBH,Feng2026GenSearcherRA,He2026MindBrushIA}, memory~\citep{He2026GEMSAM}, and self feedback~\citep{Jiang2026GenAgentST,Wang2025ImAgentAU}, but these efforts remain fragmented and do not provide a unified framework for context-centered generation.

To this end, we propose \textbf{Qwen-Image-Agent}, a unified agentic framework that integrates plan, reason, search, memory and feedback in a context-centric manner. Rather than treating user context as the final generation condition, our pipeline progressively constructs the full generation context through Context-Aware Planning and Context Grounding. Specifically, \textbf{Context-Aware Planning} operates at three levels: Information-level Planning identifies missing information and routes it to appropriate grounding strategies, Content-level Planning assembles grounded context into a detailed generation specification, and Generation-level Planning allocates context in multi-image and multi-turn scenarios. \textbf{Context Grounding} collects missing context from multiple sources, including reasoning for implicit intent inference, search for factual knowledge and visual references, memory for historical and personalized context, and feedback for iterative refinement. Overall, Qwen-Image-Agent is training-free, compatible with existing image generators, and supports both multi-image and multi-turn interaction.

Existing evaluations mainly emphasize rendering abilities~\citep{Ghosh2023GenEvalAO, Hu2024ELLAED} or isolated knowledge and reasoning abilities~\citep{Niu2025WISEAW, Zhao2025EnvisioningBT}, but fail to systematically assess the capabilities required for agentic image generation. To fill this gap, we introduce \textbf{Image-Agent-Bench (IA-Bench)}, a benchmark that evaluates four core agentic capabilities: Plan, Reason, Search, and Memory, over 17 real-world tasks, 730 test instances, and 1801 fine-grained binary checklist items. Each task is paired with a structured VLM-based evaluation protocol for reliable assessment.

Experiments on IA-Bench and prior benchmarks, including WISE-Verified~\citep{Niu2025WISEAW} and MindBench~\citep{He2026MindBrushIA}, show that Qwen-Image-Agent substantially outperforms strong agentic baselines and achieves state-of-the-art results. Ablation studies further verify the complementary benefits of different grounded contexts. Our contributions are summarized as follows:
\begin{itemize}[leftmargin=*, itemsep=2pt, topsep=2pt, parsep=0pt, partopsep=0pt]
\item We identify the \textbf{Context Gap}, i.e., the mismatch between user context and generation context as a fundamental challenge in real-world image generation. This provides a unified lens for understanding why current T2I systems fail in practical settings.
\item We propose \textbf{Qwen-Image-Agent}, a unified and context-centric framework for agentic image generation that addresses the context gap through plan, reason, search, memory and feedback.
\item We introduce \textbf{IA-Bench}, a benchmark for systematically evaluating agentic image generation along four capabilities: Plan, Reason, Search, and Memory.
\item Experiments show that Qwen-Image-Agent substantially outperforms strong agentic baselines, and achieve state-of-the-art performance on IA-Bench, Mindbench and WISE-Verified.
\end{itemize}
\section{Related Work}

\subsection{Agentic Image Generation}

Recent work extends image generation and editing with agent capabilities such as planning, reasoning, memory, search, and self feedback. Planning-based methods decompose complex intents into intermediate steps~\citep{Yao2026PhotoAgentAP}; Reasoning-based methods handle implicit user intent for more intelligent generation and editing~\citep{He2026MindBrushIA}; Search-based methods incorporate web search and image search to improve grounding in open-world scenarios~\citep{Feng2026GenSearcherRA,He2026MindBrushIA}; Memory-based methods support long-horizon interactions through persistent memory~\citep{He2026GEMSAM}; and Feedback-based methods study test-time scaling for image generation~\citep{Jiang2026GenAgentST,Wang2025ImAgentAU}. However, from the perspective of generation context, existing methods remain fragmented in how they identify, acquire, and use the context required for real-world image generation. In contrast, Qwen-Image-Agent unifies plan, reason, memory, search, and feedback within a single framework, bridging the context gap in real-world image generation.

\subsection{Benchmarks for Image Generation}

Early image generation benchmarks mainly evaluate instruction following and text--image alignment, such as GenEval~\citep{Ghosh2023GenEvalAO} for compositional attribute binding and DPGBench~\citep{Hu2024ELLAED} for dense prompt following. More recent benchmarks target harder settings that are either knowledge-driven or reasoning-driven. Knowledge-driven benchmarks, such as WISE~\citep{Niu2025WISEAW} and PhyBench~\citep{Meng2024PhyBenchAP}, evaluate grounding in domain knowledge and physical commonsense. Reasoning-driven benchmarks, such as RISEBench~\citep{Zhao2025EnvisioningBT}, test whether models can translate logical, causal, and spatio-temporal reasoning into visual outputs. Mind-Bench~\citep{He2026MindBrushIA} covers both aspects. However, existing benchmarks mainly evaluate partial agent abilities, especially reasoning or search, while largely overlooking planning and memory. To support holistic evaluation of agentic image generation, we introduce IA-Bench, which covers the full spectrum of agent capabilities with fine-grained, checklist-based evaluation.

\section{Qwen-Image-Agent Framework}
\subsection{Formulation of Image Agents}
We formalize image generation and edit as a conditional rendering problem. Given a user context $c_u=(P, I_{\mathrm{ref}})$ with prompt $P$ and optional reference images $I_{\mathrm{ref}}$, \textbf{Direct image generation} renders output image $y$ in a single forward pass, where $p_{\mathrm{gen}}$ is the image generator:

\begin{equation}
y \sim p_{\mathrm{gen}}(\cdot \mid c_u).
\end{equation}

In real-world scenarios, however, the provided user context is often incomplete for the desired visual task. We therefore distinguish \textbf{user context $c_u$} from the \textbf{generation context $c_g$}, which denotes the complete context needed for successful rendering. The earlier mentioned \textbf{context gap} is thus defined as the discrepancy between $c_u$ and $c_g$.

\textbf{Agentic image generation} addresses this gap by treating $p_{\mathrm{gen}}$ as a renderer and introducing a context-construction process to resolve the context gap. At each step $t$, the agent maintains a state $s_t$, takes an action $a_t$, and receives an observation $o_t$, forming a trajectory
\begin{equation}
\tau = \{(s_t, a_t, o_t)\}_{t=1}^{T}.
\end{equation}
The action space consists of basic operations to gather context, including \texttt{plan}, \texttt{reason}, \texttt{search}, \texttt{rewrite}, and \texttt{evaluate}. The state is defined as $s_t = (c_t, O_{t-1})$ where $c_t$ is the current context under construction, and $O_{t-1}=\{o_1,\dots,o_{t-1}\}$ is the set of accumulated intermediate results. Let $c(\tau)$ denote the final generation context induced by trajectory $\tau$. The agentic generation process is then formulated as:
\begin{equation}
p_{\mathrm{agent}}(y \mid c_u)
=
\sum_{\tau}
p(\tau \mid c_u)\,
p_{\mathrm{gen}}(y \mid c_g=c(\tau)).
\end{equation}
Under this formulation, the agent progressively builds the generation context along the trajectory before the final rendering step.

\begin{figure*}[t]
    \centering
    \includegraphics[width=\linewidth]{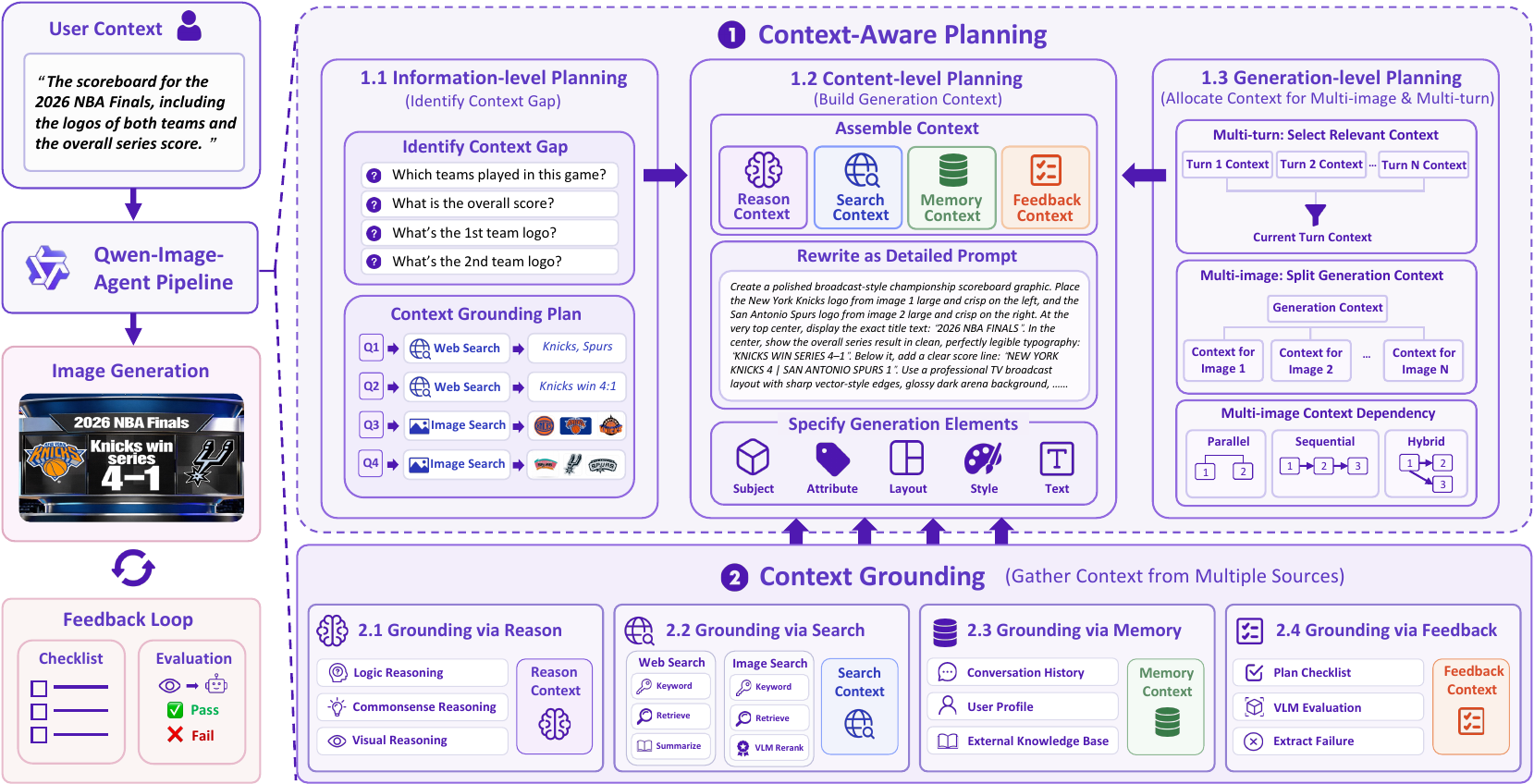}
    \caption{Overview of the Qwen-Image-Agent framework. Given a user context, the pipeline first identifies the context gap through information-level planning and gathers heterogeneous contexts. It then builds generation context through content-level planning. Qwen-Image-Agent further supports multi-turn and multi-image generation through generation-level planning.}
    \label{fig:pipeline}
\end{figure*}

\subsection{Overview of Qwen-Image-Agent}

To bridge the context gap between user context and generation context required for image generators, we propose \textbf{Qwen-Image-Agent}, a unified agentic framework that integrates planning, reasoning, search, memory and feedback in a context-centric manner. As shown in Figure~\ref{fig:pipeline}, it consists of two main modules: Context-Aware Planning and Context Grounding.

\paragraph{Context-Aware Planning} identifies missing context, plans how to obtain it, determines how it should be used for generation and how to allocate it in multi-turn and multi-image scenarios.

\paragraph{Context Grounding} gathers the missing information from multiple sources (including reason, search, memory and feedback) and organizes them in a context-centric manner.

Given a user context, the system first performs information-level planning to identify the context gap. It then grounds the missing information through reasoning, web search, and image search, producing reasoning context and search context. Together with memory context, these signals are fed into content-level planning, which builds a richer and more complete generation context for image synthesis. After an image is generated, the system evaluates the result through a feedback loop, and the newly obtained feedback context is incorporated back into content-level planning for iterative refinement. Finally, generation-level planning further extends the framework to support multi-turn and multi-image generation.

\subsection{Context-Aware Planning}

To systematically manage and utilize context throughout the generation process, we propose \textbf{Context-Aware Planning}. It operates at three levels: information-level, content-level, and generation-level.

\paragraph{Information-level planning} identifies the context gap and plans how to resolve it. Given a user context, the system first raises explicit questions to characterize the missing information required for generation. Then, it routes each questions to a suitable context grounding strategy, including reasoning, web search and image search, as detailed in Section~\ref{sec:context_grounding}.

\paragraph{Content-level planning} builds the generation context and plans the image content to be generated. Specifically, the system first assembles the context obtained during the context grounding stage, and then rewrites the user prompt into a detailed prompt that specifies key generation elements, including subject, attributes, layout, style, and textual elements.

\paragraph{Generation-level planning} allocates generation context in multi-image and multi-turn scenarios. In multi-turn settings, excessively long contexts often lead to content drift or even generation collapse. To mitigate this issue, we select relevant information from previous turns while keeping the overall context length manageable. In multi-image settings, we distribute the generation context across individual images while accounting for multi-image context dependency, including parallel, sequential, and hybrid.

\subsection{Context Grounding}~\label{sec:context_grounding}

To bridge the gap between user context and generation context, we propose \textbf{Context Grounding}, a unified module that collects context through reason, search, memory and feedback, and grounds generation with gathered context.

\paragraph{Grounding via Reason.}
User requests are often ambiguous, incomplete, or implicitly specified; therefore, generation needs to be grounded in additional context. Reasoning-based grounding addresses this issue by making implicit intents and requirements explicit. We consider three forms of reasoning: commonsense reasoning, logical reasoning, and visual reasoning. Specifically, for each question identified during Information-level Planning and assigned to reasoning, we employ a VLM to infer the corresponding answer. Together, these reasoning processes transform underspecified requests into concrete and explicit context items for downstream generation.

\paragraph{Grounding via Search.}
Some user requests depend on up-to-date factual information or IP-related visual references that cannot be inferred from the prompt alone. In such cases, we ground generation through search. For factual knowledge, we first extract search keywords from the user request, then perform web search and summarize the retrieved results into concise answers. For visual references, we retrieve candidate images from the web and employ a VLM to rank them, retaining the most relevant ones. Overall, search-based grounding enriches the request with external factual and visual context that cannot be obtained through reasoning alone.

\paragraph{Grounding via Memory.}
In multi-turn scenarios or long-horizon tasks, users may refer to knowledge or references mentioned in previous turns. In such cases, we ground generation with memory. Specifically, we incorporate the conversation history into the context and extract as well as update user profiles for long-horizon tasks. In addition, memory grounding extends to external memory sources, such as textual and visual knowledge bases. To support this, we implement a multimodal retriever that retrieves the most relevant textual and visual items from external memory and integrates them into the grounded context for generation.

\paragraph{Grounding via Feedback.}
Text-to-image models cannot directly inspect their own outputs, which often leads to discrepancies between the prompt and the generated image. In such cases, we ground generation through feedback. Specifically, after generation, we first plan a checklist of expected image attributes, and then employ a VLM to assess each generated result against this checklist. Items that fail the evaluation are converted into feedback context and combined with the previously grounded context to refine the prompt for the next round. Overall, feedback-based grounding closes the loop between generation and evaluation, enabling iterative correction toward better alignment with user intent.

\section{IA-Bench}

\begin{figure*}[t]
    \centering
    \includegraphics[width=\linewidth]{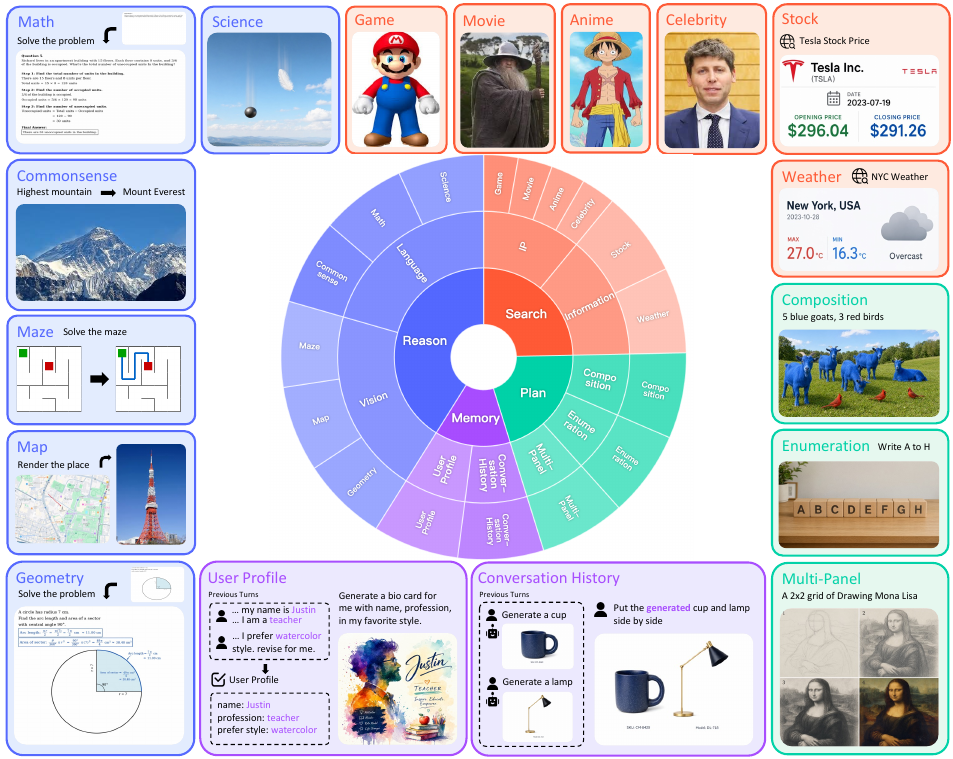}
    \caption{Overview of IA-Bench. IA-Bench covers 4 tasks, 17 subtasks, 730 instances and 1801 evaluation checklist items, providing a comprehensive evaluation of agentic image generation capabilities.}
    \label{fig:bench}
\end{figure*}

\subsection{Motivation and Overview}

Existing benchmarks for image generation mainly focus on rendering-oriented abilities, such as instruction following, visual fidelity, and aesthetic quality. However, real-world image generation often involves challenges beyond rendering alone: user requests may be underspecified, require external knowledge, demand multi-step decomposition, or depend on prior context. Addressing such requests requires models to infer implicit constraints, reason over intermediate decisions, retrieve relevant information, and maintain consistency across turns. These capabilities remain insufficiently studied in existing benchmarks, despite being particularly important for agentic image generation.

To address this gap, we introduce \textbf{Image Agent Bench (IA-Bench)}, a benchmark designed to evaluate the agentic capabilities involved in image generation. As illustrated in Figure~\ref{fig:bench}, IA-Bench covers four core capabilities: \textbf{Plan}, \textbf{Reason}, \textbf{Search}, and \textbf{Memory}. The benchmark consists of \emph{4 tasks, 17 subtasks, 730 instances and 1801 evaluation checklist items}. Together, they provide a structured evaluation framework for image generation systems across planning, reasoning, search, and memory dimensions.

\paragraph{Planning-Driven Tasks}
Planning-driven tasks evaluate whether a model can decompose a high-level goal into concrete visual arrangements and execute them in the final image. As illustrated in Figure~\ref{fig:bench}, this category includes tasks such as \textbf{Composition}, \textbf{Enumeration}, and \textbf{Multi-Panel}. These tasks require the model to explicitly organize multiple objects, satisfy counting constraints, and place visual elements into structured layouts. For example, a composition task may ask the model to place a specified number of objects with different attributes into a coherent scene, while a multi-panel task may require generating a grid of images that jointly satisfy a higher-level instruction. Such tasks emphasize deliberate planning over purely local rendering quality.

\paragraph{Reasoning-Driven Tasks}
Reasoning-driven tasks assess whether a model can infer latent constraints before generation and correctly ground the inferred result into the image. This category includes \textbf{Math}, \textbf{Science}, \textbf{Commonsense}, \textbf{Maze}, \textbf{Map}, and \textbf{Geometry}. These tasks involve three major types of reasoning: logical reasoning, commonsense reasoning, and visual reasoning. For example, a model may need to solve a math problem, infer the correct target from commonsense knowledge, or identify a valid path in a maze before rendering the final image. Unlike standard rendering tasks, success in this category depends on whether the model can first derive the correct intermediate conclusion and then faithfully express it in visual form.

\paragraph{Search-Driven Tasks}
Search-driven tasks assess whether a model can retrieve or ground external world knowledge that is not fully specified in the prompt. In IA-Bench, this category covers two major sources of knowledge: \textbf{IP}-related entities and \textbf{Information}. The \textbf{IP} branch includes tasks such as \textbf{Game}, \textbf{Movie}, \textbf{Anime}, and \textbf{Celebrity}, where the model must identify or accurately render well-known characters or people from cultural knowledge. The \textbf{Information} branch includes \textbf{Stock} and \textbf{Weather}, which require grounding up-to-date or structured real-world information into images. These tasks test whether image agents can go beyond prompt-local semantics and leverage retrieval or world knowledge to produce contextually correct outputs.

\paragraph{Memory-Driven Tasks}
Memory-driven tasks evaluate whether a model can preserve and reuse context across turns. This capability is essential for interactive image agents that must remain consistent with user preferences and prior dialogue history. IA-Bench includes \textbf{User Profile} and \textbf{Conversation History} task families. In user-profile tasks, the model must remember persistent user attributes, such as identity, profession, or preferred visual style, and incorporate them into later generations. In conversation-history tasks, the model must integrate previously generated content or earlier instructions into subsequent outputs, ensuring cross-turn consistency and correct composition. These tasks explicitly test whether the model can maintain coherent long-range context rather than treating each generation request independently.

\subsection{Benchmark Construction}

IA-Bench is constructed through careful human annotation with explicit attention to both quality and difficulty. During prompt collection, we filter out instances that can be solved by memorization or pretrained visual priors rather than the intended capability. For example, in IP-related tasks, we exclude highly iconic characters that text-to-image models can often generate correctly without external search. For each task, we further verify feasibility and minimize ambiguity in evaluation.

For checklist construction, annotators first use LLMs to generate candidates, which are then manually reviewed and refined to ensure that each item is correct and necessary. For memory-oriented tasks, we further design dynamic evaluation checklists, as the reference may be determined by images generated in earlier interaction turns rather than a static target.

\subsection{Evaluation Criterion}

To enable objective and fine-grained evaluation, we adopt a checklist-based evaluation protocol. For each test instance $i$, let $I^{i}_{\mathrm{gen}}$ denote the generated image and $\mathcal{C}^{i} = \{c^{i}_{j}\}_{j=1}^{K_i}$ denote its associated checklist, where each item corresponds to a required visual condition. We use a VLM to determine whether the generated image satisfies each checklist item. We report two complementary metrics:

\paragraph{Pass Rate (PR)}
Pass rate measures strict task success. An instance is considered successful only when all checklist items are satisfied:
\[
\mathrm{PR} = \frac{1}{N}\sum_{i=1}^{N}\prod_{j=1}^{K_i} \mathrm{VLM}(I^{i}_{\mathrm{gen}}, c^{i}_{j}).
\]

\paragraph{Checklist Accuracy (CA)}
Checklist accuracy measures the average proportion of checklist items satisfied by the generated image:
\[
\mathrm{CA} = \frac{1}{N}\sum_{i=1}^{N}\left(\frac{1}{K_i}\sum_{j=1}^{K_i} \mathrm{VLM}(I^{i}_{\mathrm{gen}}, c^{i}_{j})\right).
\]

Pass rate reflects strict end-to-end completion under all required constraints, whereas Checklist accuracy captures partial compliance in multi-constraint generation settings. Together, they characterize both holistic completion and partial fulfillment.

\paragraph{Image Agent score (IA-score)}
To summarize overall agent performance across different capability dimensions, we further report \textbf{IA-score}, a weighted aggregate score over the four core dimensions in IA-Bench: \textit{Plan}, \textit{Reason}, \textit{Search}, and \textit{Memory}. Specifically, IA-score is defined as
\[
\mathrm{IA\text{-}score} = 0.3 \times \mathrm{Plan} + 0.3 \times \mathrm{Reason} + 0.3 \times \mathrm{Search} + 0.1 \times \mathrm{Memory}.
\]
Here, \textit{Plan}, \textit{Reason}, \textit{Search}, and \textit{Memory} denote the micro average evaluation scores for their respective dimensions. We assign higher weights to \textit{Plan}, \textit{Reason}, and \textit{Search}, as these dimensions capture the core capabilities required for real-world image-agent tasks, while \textit{Memory} is included as a complementary factor for measuring cross-step consistency and context retention.

\section{Experiments}

\subsection{Experimental Settings}

\paragraph{Benchmarks}
To comprehensively evaluate the ability of existing methods, we consider three benchmarks. First, our proposed \textbf{IA-Bench}, which measures four core agentic capabilities including Plan, Reason, Search and Memory. Second, \textbf{WISE-Verified}~\citep{Niu2025WISEAW}, a human-verified version of WISE, assesses semantic understanding and world knowledge in image generation models. Finally, \textbf{MindBench}~\citep{He2026MindBrushIA} evaluates the use of dynamic external knowledge and multi-step reasoning.
\paragraph{Baselines} We compare Qwen-Image-Agent against \textbf{proprietary models}, including GPT-Image-1~\citep{openai2024gptimage1}, GPT-Image-1.5~\citep{openai2025gptimage15}, Nano Banana~\citep{deepmind2024geminiimage25}, Nano Banana Pro~\citep{deepmind2024geminiimage}, FLUX.2-pro~\citep{blackforestlabs2026flux2pro}, FLUX.2-max~\citep{blackforestlabs2026flux2max}, Seedream-5.0-Lite~\citep{bytedance_seedream5_lite_2025} and Qwen-Image-2.0~\citep{Zhao2026QwenImage20TR}, as well as state-of-the-art \textbf{open-source models} including Stable Diffusion series~\citep{podell2023sdxl, rombach2022high, sd3Medium, sd35Medium, sd35Large}, FLUX series~\citep{flux2024, labs2025flux1kontextflowmatching, blackforestlabs2026flux2pro}, Janus series~\citep{wu2024janus, chen2025janus}, Z-Image~\citep{cai2025z}, Qwen-Image~\citep{wu2025qwen}, and \textbf{unified generation models} including UniWorld-V1~\citep{lin2025uniworld}, Bagel~\citep{deng2025bagel}, Echo-4o~\citep{ye2025echo}, and DraCo~\citep{jiang2025draco}. We also include a wide range of \textbf{agentic generation models} including GEMS~\citep{He2026GEMSAM}, MindBrush~\citep{He2026MindBrushIA}, GenSearcher~\citep{Feng2026GenSearcherRA} and SCOPE~\citep{Ren2026SCOPESD}. All baselines are evaluated in their default settings.

\paragraph{Implementation Details}
We employ Qwen-Image-2.0 as the image generation and edit backbone, and GPT-5.5-0424 as the MLLM backbone. Regarding search tools, we utilize Google Search API for web search and image search. We set the limit of text search to 5, and the limit of image search to 5. We further utilize Jina API to process visited web pages. To ensure a fair comparison, all agentic generation baselines are evaluated under the same experimental setting, using GPT-5.5-0424 as the MLLM backbone and Qwen-Image-2.0 as the image generation and edit backbone. For the feedback loop, we allow up to 3 feedback attempts on IA-Bench, while disabling the feedback loop on WISE-Verified and MindBench to enable direct comparison with non-agentic methods. In IA-Bench, for the baselines without multiturn abilities, we append the previous turn information as prompt for testing.

\subsection{Quantitative Results}
\begin{table*}[t]
\centering
\small
\setlength{\tabcolsep}{3pt}
\renewcommand{\arraystretch}{1.05}
\resizebox{\textwidth}{!}{%
\begin{tabular}{lccccccccc}
\toprule
\multirow{2}{*}{\textbf{Model Name}} 
& \multicolumn{4}{c}{\textbf{Checklist Accuracy (\%)}} 
& \multicolumn{4}{c}{\textbf{Pass Rate (\%)}} 
& \multirow{2}{*}{\textbf{IA-score}} \\
\cmidrule(lr){2-5} \cmidrule(lr){6-9}
& \textbf{Plan} & \textbf{Reason} & \textbf{Search} & \textbf{Memory}
& \textbf{Plan} & \textbf{Reason} & \textbf{Search} & \textbf{Memory}
&  \\
\midrule

\rowcolor{gray!8}
\multicolumn{10}{l}{\textit{Closed-source Image Generation Models}} \\
GPT-Image-1.5~\citep{openai2025gptimage15}
& 55.1 & 55.6 & 55.2 & 87.6
& 23.3 & 36.7 & 35.0 & 72.0
& 35.7 \\
Nano Banana~\citep{deepmind2024geminiimage25}
& 68.0 & 63.9 & 61.7 & 60.2
& 42.0 & 43.3 & 42.2 & 48.0
& 43.1 \\
Nano Banana Pro~\citep{deepmind2024geminiimage}
& 60.8 & 66.2 & 68.3 & 72.0
& 32.7 & 44.3 & 47.8 & 52.0
& 42.6 \\
Seedream-5.0-Lite~\citep{bytedance_seedream5_lite_2025}
& 71.3 & 58.3 & 50.1 & 66.4
& 46.0 & 37.0 & 21.1 & 48.0
& 36.0 \\
Qwen-Image-2.0~\citep{Zhao2026QwenImage20TR}
& 50.0 & 48.2 & 38.0 & 51.8
& 20.0 & 27.7 & 6.7 & 11.0
& 17.4 \\

\midrule
\rowcolor{gray!8}
\multicolumn{10}{l}{\textit{Open-source Image Generation Models}} \\
SD-3.5-medium~\citep{sd35Medium}
& 15.6 & 6.9 & 20.4 & 5.9
& 0.0 & 4.0 & 3.3 & 0.0
& 2.2 \\
SD-3.5-large~\citep{sd35Large}
& 19.0 & 9.2 & 24.2 & 10.5
& 0.0 & 5.7 & 6.1 & 1.0
& 3.6 \\
FLUX.2-dev~\citep{flux2024}
& 29.4 & 23.2 & 33.1 & 52.9
& 5.3 & 15.0 & 9.4 & 11.0
& 10.0 \\
Bagel~\citep{deng2025bagel}
& 20.0 & 12.6 & 15.1 & 4.7
& 0.7 & 4.0 & 0.6 & 0.0
& 1.6 \\
Bagel w/ CoT~\citep{deng2025bagel}
& 22.6 & 26.7 & 12.8 & 5.9
& 2.0 & 19.0 & 0.6 & 0.0
& 6.5 \\
Echo-4o~\citep{ye2025echo}
& 22.1 & 11.6 & 17.1 & 7.9
& 0.7 & 4.0 & 0.6 & 0.0
& 1.6 \\
Echo-4o w/ CoT~\citep{ye2025echo}
& 17.4 & 10.1 & 9.3 & 7.6
& 0.0 & 4.0 & 0.6 & 0.0
& 1.4 \\
Qwen-Image~\citep{wu2025qwen}
& 30.0 & 28.2 & 35.1 & 41.1
& 4.7 & 17.7 & 6.1 & 9.0
& 9.4 \\

\midrule
\rowcolor{gray!8}
\multicolumn{10}{l}{\textit{Agentic Image Generation Models}} \\
GenSearcher~\citep{Feng2026GenSearcherRA}
& 37.0 & 30.1 & 46.5 & 46.6
& 9.3 & 20.3 & 24.4 & 11.0
& 17.3 \\
GEMS~\citep{He2026GEMSAM}
& 70.6 & 28.4 & 49.4 & 52.6
& 41.3 & 18.3 & 18.9 & 13.0
& 24.9 \\
MindBrush~\citep{He2026MindBrushIA}
& 56.1 & 51.8 & 53.6 & 53.1
& 28.0 & 32.7 & 35.6 & 13.0
& 30.2 \\
SCOPE~\citep{Ren2026SCOPESD}
& 73.3 & 45.2 & 44.4 & 45.2
& 46.7 & 30.0 & 23.3 & 9.0
& 30.9 \\
\rowcolor{blue!8}
\textbf{Qwen-Image-Agent}
& 72.9 & 65.5 & 67.6 & 73.6
& 45.3 & 43.7 & 46.1 & 49.0
& {\color{blue!70!black}\textbf{45.4}} \\
\bottomrule
\end{tabular}
}
\caption{Results on IA-Bench. We report checklist accuracy, pass rate, and the overall IA-score, all measured in percentage (\%). For all metrics, higher values indicate better performance.
}
\label{tab:ia_bench}
\end{table*}

\begin{table*}[t]
\centering
\scriptsize
\setlength{\tabcolsep}{3pt}
\renewcommand{\arraystretch}{1.05}
\resizebox{\textwidth}{!}{%
\begin{tabular}{l|cccccc|c}
\toprule
\textbf{Model Name} & \textbf{Culture} & \textbf{Time} & \textbf{Space} & \textbf{Biology} & \textbf{Physics} & \textbf{Chemistry} & \textbf{Overall} \\
\midrule
Nano Banana Pro~\citep{deepmind2024geminiimage} & {0.8975} & {0.8167} & {0.9333} & {0.8167} & {0.8667} & {0.8750} & {0.8760} \\
GPT-Image-1.5~\citep{openai2025gptimage15} & 0.8900 & 0.6917 & 0.8833 & 0.8000 & 0.7583 & 0.7750 & 0.8250 \\
Qwen-Image-2.0~\citep{Zhao2026QwenImage20TR} & 0.8219 & 0.6500 & 0.8992 & 0.7917 & 0.8000 & 0.7479 & 0.7954 \\
Bagel (w/ CoT)~\citep{deng2025bagel} & 0.7800 & 0.6333 & 0.5667 & 0.3750 & 0.5500 & 0.5083 & 0.6280 \\
Bagel~\citep{deng2025bagel} & 0.4125 & 0.3500 & 0.3083 & 0.2000 & 0.4417 & 0.2583 & 0.3520 \\
Janus-Pro-7B~\citep{chen2025janus} & 0.3700 & 0.3500 & 0.2833 & 0.2833 & 0.4000 & 0.2333 & 0.3340 \\
Janus-Pro-1B~\citep{chen2025janus} & 0.3050 & 0.2333 & 0.2333 & 0.2167 & 0.3083 & 0.2000 & 0.2650 \\
Janus-1.3B~\citep{wu2024janus} & 0.3175 & 0.2833 & 0.1833 & 0.2250 & 0.3417 & 0.1833 & 0.2730 \\
FLUX.2-dev~\citep{blackforestlabs2026flux2pro} & 0.6650 & 0.5667 & 0.6583 & 0.3667 & 0.5250 & 0.3750 & 0.5650 \\
FLUX.2-klein-9B~\citep{blackforestlabs2026flux2pro} & 0.4900 & 0.3917 & 0.5500 & 0.3833 & 0.4833 & 0.2250 & 0.4400 \\
FLUX.2-klein-4B~\citep{blackforestlabs2026flux2pro} & 0.4400 & 0.3667 & 0.4667 & 0.3167 & 0.3917 & 0.3333 & 0.4010 \\
FLUX.1-dev~\citep{flux2024} & 0.5225 & 0.4000 & 0.5333 & 0.1750 & 0.3750 & 0.2417 & 0.4160 \\
FLUX.1-schnell~\citep{flux2024} & 0.4650 & 0.3250 & 0.4667 & 0.2083 & 0.3833 & 0.1000 & 0.3640 \\
SD-3.5-large~\citep{sd35Large} & 0.4900 & 0.4083 & 0.4417 & 0.3000 & 0.3750 & 0.2083 & 0.4040 \\
SD-3.5-medium~\citep{sd35Medium} & 0.4825 & 0.3750 & 0.3750 & 0.1833 & 0.3917 & 0.2000 & 0.3760 \\
SD-3-medium~\citep{sd3Medium} & 0.4700 & 0.4083 & 0.4000 & 0.2000 & 0.3750 & 0.2583 & 0.3850 \\
SD-XL-0.9~\citep{podell2023sdxl} & 0.4925 & 0.3667 & 0.2417 & 0.2667 & 0.3333 & 0.1833 & 0.3640 \\
SD-1.5~\citep{rombach2022high} & 0.4450 & 0.3083 & 0.2083 & 0.2083 & 0.2167 & 0.1500 & 0.3090 \\
Qwen-Image~\citep{wu2025qwen} & 0.6275 & 0.5250 & 0.5583 & 0.3417 & 0.4833 & 0.2500 & 0.5100 \\
Qwen-Image-2512~\citep{wu2025qwen} & 0.5950 & 0.4750 & 0.6000 & 0.3500 & 0.4917 & 0.2583 & 0.4990 \\
UniWorld-V1~\citep{lin2025uniworld} & 0.5150 & 0.4917 & 0.5500 & 0.2250 & 0.4000 & 0.1667 & 0.4260 \\
Z-Image~\citep{cai2025z} & 0.5475 & 0.4667 & 0.5083 & 0.3250 & 0.4750 & 0.1750 & 0.4530 \\
\rowcolor{blue!10}
\textbf{Qwen-Image-Agent} & \textbf{0.9200} & \textbf{0.9167} & \textbf{0.9333} & \textbf{0.8333} & \textbf{0.8667} & \textbf{0.9000} & \textbf{0.9020} \\
\bottomrule
\end{tabular}
}
\caption{Results on WISE-Verified. Best results are shown in bold.}
\label{tab:wise}
\end{table*}
\begin{table*}[t]
\centering
\scriptsize
\setlength{\tabcolsep}{3pt}
\renewcommand{\arraystretch}{1.05}
\resizebox{\textwidth}{!}{%
\begin{tabular}{l|ccccc|ccccc|c}
\toprule
\multirow{2}{*}{\textbf{Model Name}}
& \multicolumn{5}{c|}{\textbf{Knowledge-Driven}} 
& \multicolumn{5}{c|}{\textbf{Reasoning-Driven}} 
& \multirow{2}{*}{\textbf{Overall}} \\
\cmidrule(lr){2-6} \cmidrule(lr){7-11}
& SE & Wth & MC & IP & WK & SL & Poem & LifeR & GU & Math &  \\
\midrule
GPT-Image-1~\citep{openai2024gptimage1}        & 0.32 & 0.06 & 0.22 & 0.02 & 0.16 & 0.32 & 0.10 & 0.24 & 0.10 & 0.12 & 0.17 \\
GPT-Image-1.5~\citep{openai2025gptimage15}     & 0.36 & 0.18 & 0.22 & 0.04 & 0.30 & 0.34 & 0.08 & \textbf{0.34} & 0.10 & 0.02 & 0.21 \\
FLUX.2-pro~\citep{blackforestlabs2026flux2pro} & 0.38 & 0.12 & 0.08 & 0.00 & 0.20 & 0.44 & 0.64 & 0.18 & 0.04 & 0.02 & 0.21 \\
FLUX.2-max~\citep{blackforestlabs2026flux2max} & 0.44 & 0.12 & 0.10 & 0.04 & \underline{0.38} & 0.40 & 0.50 & 0.20 & 0.02 & 0.06 & 0.23 \\
Nano Banana~\citep{deepmind2024geminiimage25}  & 0.30 & 0.10 & 0.12 & 0.00 & 0.30 & 0.32 & 0.36 & 0.20 & 0.04 & 0.08 & 0.18 \\
Nano Banana Pro~\citep{deepmind2024geminiimage} & \underline{0.50} & \textbf{0.36} & \underline{0.40} & \textbf{0.16} & \textbf{0.56} & \textbf{0.62} & \underline{0.68} & \underline{0.30} & \textbf{0.16} & \textbf{0.46} & \underline{0.41} \\
\midrule
SDXL~\citep{podell2023sdxl}                     & 0.04 & 0.00 & 0.04 & 0.00 & 0.00 & 0.00 & 0.00 & - & - & - & 0.01 \\
SD-3.5-medium~\citep{sd35Medium}               & 0.02 & 0.00 & 0.00 & 0.00 & 0.02 & 0.00 & 0.00 & - & - & - & 0.01 \\
SD-3.5-large~\citep{sd35Large}                 & 0.04 & 0.00 & 0.02 & 0.00 & 0.02 & 0.00 & 0.06 & - & - & - & 0.01 \\
FLUX.1-dev~\citep{flux2024}                    & 0.04 & 0.00 & 0.00 & 0.00 & 0.02 & 0.02 & 0.04 & - & - & - & 0.02 \\
FLUX.1-kontext~\citep{labs2025flux1kontextflowmatching} & 0.02 & 0.00 & 0.00 & 0.00 & 0.02 & 0.00 & 0.00 & - & - & - & 0.01 \\
FLUX.1-krea~\citep{flux2024}                   & 0.04 & 0.00 & 0.04 & 0.00 & 0.02 & 0.00 & 0.02 & - & - & - & 0.02 \\
Bagel~\citep{deng2025bagel}                    & 0.02 & 0.00 & 0.00 & 0.00 & 0.00 & 0.02 & 0.02 & 0.02 & 0.00 & 0.08 & 0.02 \\
Echo-4o~\citep{ye2025echo}                     & 0.04 & 0.00 & 0.00 & 0.00 & 0.00 & 0.02 & 0.06 & 0.02 & 0.02 & 0.02 & 0.02 \\
DraCo~\citep{jiang2025draco}                   & 0.02 & 0.00 & 0.02 & 0.00 & 0.00 & 0.02 & 0.02 & 0.04 & 0.02 & 0.06 & 0.02 \\
Z-Image~\citep{cai2025z}                       & 0.02 & 0.00 & 0.08 & 0.02 & 0.00 & 0.00 & 0.00 & - & - & - & 0.02 \\
Qwen-Image~\citep{wu2025qwen}                  & 0.08 & 0.00 & 0.04 & 0.00 & 0.00 & 0.04 & 0.00 & 0.04 & 0.00 & 0.00 & 0.02 \\
Qwen-Image-2.0~\citep{Zhao2026QwenImage20TR}   & 0.19 & 0.24 & 0.23 & 0.04 & 0.12 & 0.42 & 0.58 & 0.12 & 0.02 & 0.28 & 0.23 \\
\midrule
\rowcolor{blue!10}
\textbf{Qwen-Image-Agent} & \textbf{0.60} & \underline{0.28} & \textbf{0.70} & \textbf{0.16} & 0.28 & \underline{0.58} & \textbf{0.82} & 0.24 & \textbf{0.20} & \underline{0.34} & \textbf{0.42} \\
\bottomrule
\end{tabular}
}
\caption{Results on MindBench. Best results are in bold and the second best ones are underlined.}
\label{tab:mindbench}
\end{table*}

We present the quantitative results on IA-Bench in Table~\ref{tab:ia_bench}. As shown, Qwen-Image-Agent achieves the highest IA-score, outperforming strong closed-source baselines such as Nano Banana Pro and GPT-Image-1.5. Compared with the direct generation baseline Qwen-Image-2.0, our agentic framework improves the Q-score substantially, \textbf{from 17.4 to 45.4}. In comparison with other agentic image generation methods, Qwen-Image-Agent achieves strong performance across the Plan, Reason, and Search dimensions, which we attribute to its unified, context-centered framework. More importantly, it shows a particularly large improvement in the Memory dimension, which hightlights its practical value in real-world, multi-turn image generation scenarios.

From the overall comparison, we observe that agentic generation models consistently outperform direct generation models on core agentic capabilities such as Plan, Reason, and Search. At the same time, closed-source models still maintain a noticeable advantage in Memory compared with agentic methods. These findings suggest that IA-Bench is a valid and informative benchmark for evaluating image agents, while also shedding light on promising directions for future research in agentic image generation.

Moreover, Qwen-Image-Agent delivers outstanding performance on both WISE-Verified, which emphasizes world knowledge, and MindBench, which focuses on complex reasoning and the use of external knowledge. As shown in Table~\ref{tab:wise}, on WISE-Verified, Qwen-Image-Agent achieves state-of-the-art performance, surpassing the previous SOTA model, Nano Banana Pro. The results on MindBench are reported in Table~\ref{tab:mindbench}, where Qwen-Image-Agent also sets a new state of the art. In particular, compared with the direct generation baseline Qwen-Image-2.0, our agentic framework improves performance by 82.6\%. These results further demonstrate the practical effectiveness and generalizability of our proposed agentic framework across diverse image generation tasks.

\subsection{Qualitative Results}
\begin{figure*}[t]
    \centering
    \includegraphics[width=\linewidth]{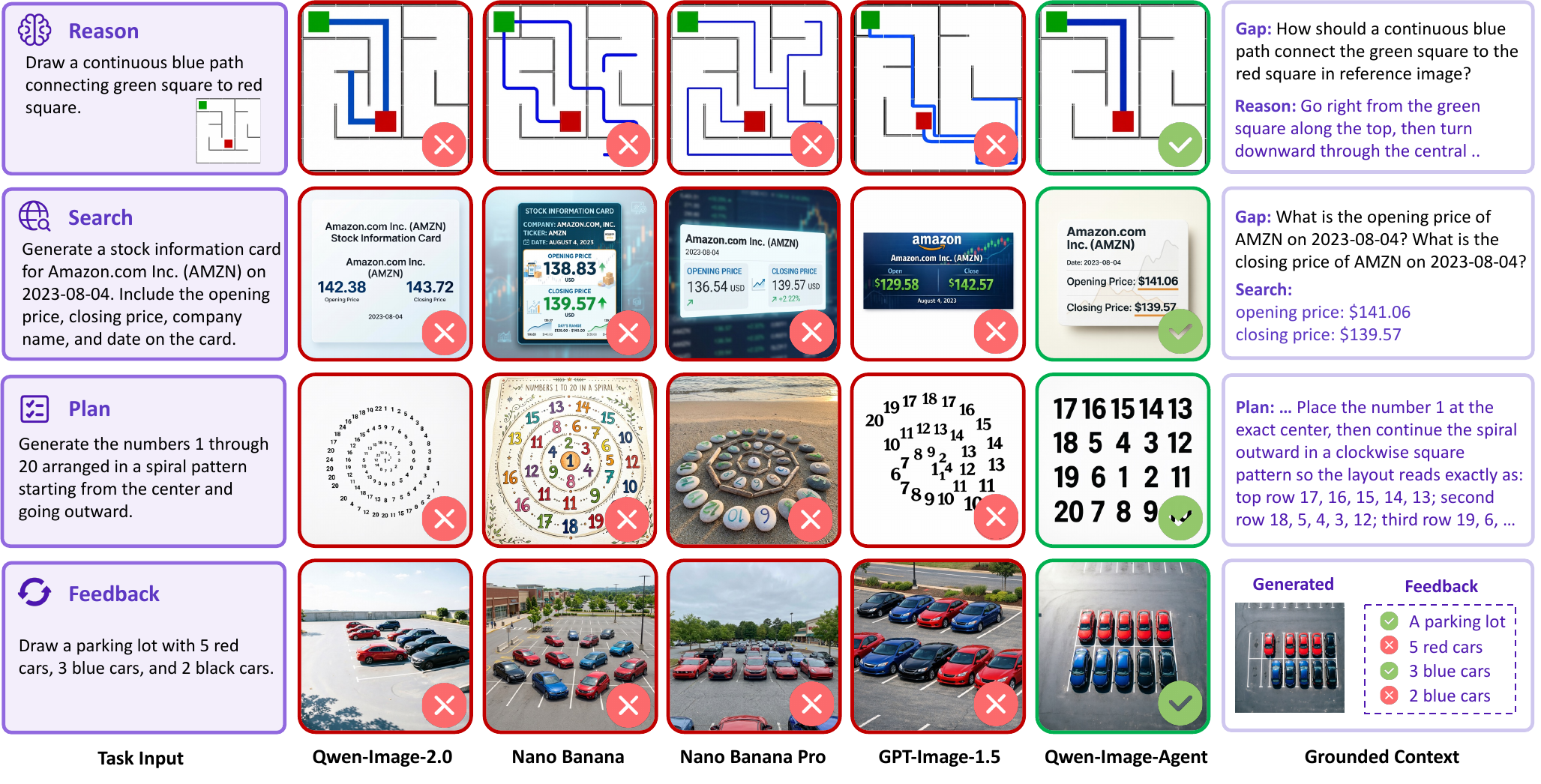}
    \caption{Qualitative Comparison of different models on IA-Bench, which demonstrates different capabilities of Qwen-Image-Agent, including Plan, Reason, Search, and Feedback.}
    \label{fig:qualitative}
\end{figure*}

Figure~\ref{fig:qualitative} presents a qualitative comparison between Qwen-Image-Agent and strong baselines, including Qwen-Image-2.0, NanoBanana, NanoBanana Pro, and GPT-Image-1.5. Although Qwen-Image-Agent is built upon Qwen-Image-2.0, it substantially improves generation quality on complex real-world tasks by bridging the context gap through our agentic pipeline. Instead of directly treating the user request as the final generation condition, Qwen-Image-Agent progressively transforms incomplete user context into sufficient generation context for image synthesis.

As shown in the figure, Qwen-Image-Agent can infer the correct maze trajectory in the reasoning case, retrieve accurate stock information in the search case, generate the specified spiral layout in the planning case, and verify object attributes and composition in the feedback case. In contrast, existing baselines often fail when the required context is implicit, missing, or needs to be grounded before generation. These examples demonstrate the effectiveness of our proposed pipeline and highlight the importance of addressing the context gap in real-world image generation.

\begin{table*}[t]
\centering
\fontsize{8.5pt}{10pt}\selectfont
\setlength{\tabcolsep}{3.5pt}
\renewcommand{\arraystretch}{1.05}
\begin{tabular}{ccc|cccc|c}
\toprule
\multirow{2}{*}{\textbf{Framework}} & \multirow{2}{*}{\textbf{MLLM backbone}} & \multirow{2}{*}{\textbf{Gen. backbone}}
& \multicolumn{4}{c|}{\textbf{Pass Rate (\%)}} 
& \multirow{2}{*}{\textbf{IA-score}} \\
\cmidrule(lr){4-7}
& & & \textbf{Plan} & \textbf{Reason} & \textbf{Search} & \textbf{Memory} & \\

\midrule
Qwen-Image-Agent & GPT-55 & Qwen-Image-2.0
& 45.3 & 43.7 & 46.1 & 49.0 & 45.4 \\
w/o Reason & GPT-55 & Qwen-Image-2.0
& \textcolor{green!60!black}{24.7} & \textcolor{green!60!black}{29.7} & 46.1 & 49.0 & 35.1 \\
w/o Search & GPT-55 & Qwen-Image-2.0
& 46.0 & 44.3 & \textcolor{green!60!black}{7.8} & 49.0 & 34.3 \\
w/o Memory & GPT-55 & Qwen-Image-2.0
& 45.3 & 43.7 & 46.1 & \textcolor{green!60!black}{0.0} & 40.5 \\
w/o Feedback & GPT-55 & Qwen-Image-2.0
& 40.0 & 41.3 & 42.8 & 49.0 & 42.1 \\

\midrule
Qwen-Image-Agent & GPT-55 & Qwen-Image
& \textcolor{green!60!black}{19.3} & \textcolor{green!60!black}{30.7} & 31.1 & 40.0 & 28.3 \\

Qwen-Image-Agent & Qwen & Qwen-Image-2.0
& \textcolor{green!60!black}{24.7} & 41.7 & \textcolor{green!60!black}{19.4} & \textcolor{green!60!black}{21.0} & 27.8 \\
\bottomrule
\end{tabular}
\caption{Ablation study on Grounded Context, MLLM Backbone, and Generation Backbone, conducted on IA-Bench using Pass Rate as metric. Metrics with significant decreases are marked in green}
\label{tab:ablation}
\end{table*}
\subsection{Ablation Study}
\paragraph{Ablations on Grounded Context}
To validate the effectiveness of grounded contexts in Qwen-Image-Agent, we conduct comprehensive ablation studies on different types of grounded contexts, including \textit{Reason Context}, \textit{Search Context}, \textit{Memory Context}, and \textit{Feedback Context}, using IA-Bench evaluation protocols. As shown in Table~\ref{tab:ablation}, removing any grounded context leads to a clear drop in its corresponding evaluation dimension. This not only verifies the effectiveness of our context design, but also supports the validity of IA-Bench, as each dimension is sensitive to the capability it is intended to measure. We also observe that removing \textit{Reason Context} degrades both Reason and Plan. This is because some implicit user requirements, such as enumeration, are resolved during reasoning and then reflected in planning. By contrast, removing \textit{Feedback Context} causes a relatively smaller drop, which we attribute to the strong rendering accuracy of Qwen-Image-2.0. Overall, these results support our main claim that bridging the context gap greatly improves real-world image generation.

\paragraph{Ablations on MLLM Backbone}
To study the impact of the MLLM backbone, we conduct ablations on the backbone choice. By default, we use GPT-5.5-0424 as the MLLM backbone. In the ablation setting, we replace it with Qwen-Plus as the LLM backbone and Qwen-VL-Max as the VLM backbone. As shown in Table~\ref{tab:ablation}, replacing the default MLLM backbone causes substantial degradation across most metrics, showing that MLLM intelligence is critical to the overall system. In particular, it is important for layout-aware planning, keyword generation and information integration in search, and relevant context selection in memory.

\paragraph{Ablations on Generation Backbone}
To investigate the impact of image renderers under a fixed generation context, we conduct ablations on the image generation and editing backbones. By default, we use Qwen-Image-2.0 as the image generation and edit backbone. In the ablation setting, we use Qwen-Image as the generation backbone and Qwen-Image-Edit as the edit backbone. Table~\ref{tab:ablation} shows that changing the generation backbone leads to consistent performance drops across all metrics. This suggests that strong generation and editing capability is also necessary for the full system. Even with a complete prompt and correct planning, some tasks remain difficult due to renderer limitations, such as counted composition, visually grounded reasoning, and accurate visual reference following.

\subsection{Discussion}
Through our experiments, we identify and summarize several important challenges and common failure modes in agentic image generation. These findings explain where current systems still struggle, shedding light on the main bottlenecks beyond direct image rendering.

\paragraph{Unidentified Context Gaps}
One of the central challenges in agentic image generation is identifying the gap between user context and generation context. Still, in some user cases, the context gap remains too implicit to be reliably identified, such as when the model must infer a historical event from a specific date and location stated in prompt. We find that such failures cannot be addressed by a stronger Generation backbone, since the bottleneck lies before rendering. Instead, they largely depend on the intelligence of the MLLM backbone for recognizing the missing context. Thus, we adopt a stronger MLLM backbone and substantially improves the overall system.

\paragraph{Ambiguous Boundary between Reason and Search}
The boundary between reasoning and search is often unclear. Some facts can be solved either by parametric knowledge or by external retrieval, depending on the capability boundary of the MLLM backbone. In our framework, we treat commonsense facts as solvable by internal reasoning, and define two categories that require explicit search: Precise Facts, which demand exact factual accuracy such as specific numbers, dates, and names, and Dynamic Facts, which change over time. We find that this definition helps decouple reasoning from search in a principled way, and our ablation results further support the effectiveness of this design.

\paragraph{Excessive Image Search}
Although image search provides useful visual grounding, excessive image search may hurt final generation quality: (1) Current editing models are generally less robust than direct generation models, and multi-reference editing is often more brittle than single-reference conditioning. (2) Irrelevant or weakly related reference images introduce harmful visual bias and degrade the final output. In particular, we observe that some agentic baselines, such as GenSearcher, tend to overuse image retrieval, which introduces distracting visual references and degrades the output. This issue is closely related to the IP capability of the Generation backbone. We therefore adapt the boundary of image search to the capability of the underlying generator. In our case, Qwen-Image-2.0 still lags behind the strongest models on IP-related tasks, we thus explicitly invoke image search for clear IP reference needs, while enforcing relatively strict constraints to avoid unnecessary visual retrieval.

\paragraph{Context Explosion in Multiturn Generation}
A major challenge in multiturn generation is context explosion, especially the rapid growth of image-token context. Across multiple turns, the system may need to process user-provided image references, previously generated images, and images retrieved from search, all of which consume substantial visual context. We observe cases where such accumulated multimodal context already exceeds the token limits of strong baselines such as Nano Banana and Nano Banana Pro, leading to generation failure. To mitigate this issue, our system performs relevance-based context selection rather than naively retaining all historical inputs. This substantially alleviates context explosion and is critical for maintaining stable performance in long-horizon multiturn interactions.

\paragraph{Weak Feedback Supervision}
We also observe that the gains from feedback are relatively limited in our current setting. We attribute this to two main reasons. (1) First, our current feedback mechanism is implemented as a prompt-based feedback loop at the generation stage. In future work, we plan to extend feedback beyond post-hoc critique, so that it can also supervise context-gap identification and context grounding earlier in the pipeline. (2) Second, because we target general-purpose scenarios, we currently rely on VLM-generated feedback checklists as a generic feedback signal. In many real applications, however, one can introduce more explicit and task-specific supervision, such as predefined downstream metrics, generation quality criteria, or learned reward models. Such signals would provide clearer and more targeted feedback, and could potentially support stronger test-time scaling.

\paragraph{High Latency and Cost}
The full agentic pipeline inevitably introduces higher latency and cost than direct generation, since it may involve plan, reason, search, context integration, generation, and feedback loop. To mitigate this, we organize both information-level planning and generation-level planning with DAG-based execution, enabling as much parallelism as possible. Still, the overall pipeline remains substantially more expensive than one-shot generation. This highlights the need for more efficient agentic pipelines, potentially through training-based optimization or better tool-use policies.
\section{Conclusion}
In this work, we identify the context gap as a central challenge in real-world image generation. To address it, we propose \textbf{Qwen-Image-Agent}, a unified agentic framework that integrates plan, reason, search, memory and feedback in a context-centric manner. We further introduce \textbf{IA-Bench}, a benchmark for systematically evaluating four core capabilities of agentic image generation: Plan, Reason, Search, and Memory. Overall, our work highlights a shift from direct image generation to agentic image generation, and provides a unified context-centric perspective for understanding this transition.
We hope our work offers practical guidance for building future image agents that can go beyond direct prompt rendering and better address real-world user needs.

\clearpage
\bibliography{colm2024_conference}

@article{Ye2026AgentBH,
  title={Agent Banana: High-Fidelity Image Editing with Agentic Thinking and Tooling},
  author={Ruijie Ye and Jiayi Zhang and Zhuoxin Liu and Zihao Zhu and Siyuan Yang and Li Li and Tianfu Fu and Franck Dernoncourt and Yue Zhao and Jiacheng Zhu and Ryan A. Rossi and Wenhao Chai and Zhengzhong Tu},
  journal={ArXiv},
  year={2026},
  volume={abs/2602.09084},
  url={https://api.semanticscholar.org/CorpusID:285462350}
}

@inproceedings{He2026GEMSAM,
  title={GEMS: Agent-Native Multimodal Generation with Memory and Skills},
  author={Zefeng He and Siyuan Huang and Xiaoye Qu and Yafu Li and Tong Zhu and Yu Cheng and Yang Yang},
  year={2026},
  url={https://api.semanticscholar.org/CorpusID:286974454}
}

@article{Jiang2026GenAgentST,
  title={GenAgent: Scaling Text-to-Image Generation via Agentic Multimodal Reasoning},
  author={Kaixun Jiang and Yuzheng Wang and Junjie Zhou and Pandeng Li and Zhihang Liu and Chen-Wei Xie and Zhaoyu Chen and Yun Zheng and Wenqiang Zhang},
  journal={ArXiv},
  year={2026},
  volume={abs/2601.18543},
  url={https://api.semanticscholar.org/CorpusID:285050929}
}

@inproceedings{Feng2026GenSearcherRA,
  title={Gen-Searcher: Reinforcing Agentic Search for Image Generation},
  author={Kaituo Feng and Manyuan Zhang and Shuang Chen and Yunlong Lin and Kaixuan Fan and Yilei Jiang and Hongyu Li and Dian Zheng and Chenyang Wang and Xiangyu Yue},
  year={2026},
  url={https://api.semanticscholar.org/CorpusID:286975158}
}

@article{Wang2025ImAgentAU,
  title={ImAgent: A Unified Multimodal Agent Framework for Test-Time Scalable Image Generation},
  author={Kaishen Wang and Ruibo Chen and Tong Zheng and Heng Huang},
  journal={ArXiv},
  year={2025},
  volume={abs/2511.11483},
  url={https://api.semanticscholar.org/CorpusID:283055363}
}

@article{He2026MindBrushIA,
  title={Mind-Brush: Integrating Agentic Cognitive Search and Reasoning into Image Generation},
  author={Jun He and Junyan Ye and Zilong Huang and Dongzhi Jiang and Chenjue Zhang and Leqi Zhu and Renrui Zhang and Xiang Zhang and Weijia Li},
  journal={ArXiv},
  year={2026},
  volume={abs/2602.01756},
  url={https://api.semanticscholar.org/CorpusID:285269721}
}

@article{Yao2026PhotoAgentAP,
  title={PhotoAgent: Agentic Photo Editing with Exploratory Visual Aesthetic Planning},
  author={Mingde Yao and Zhiyuan You and King-Man Tam and Menglu Wang and Tianfan Xue},
  journal={ArXiv},
  year={2026},
  volume={abs/2602.22809},
  url={https://api.semanticscholar.org/CorpusID:286082495}
}

@article{Ghosh2023GenEvalAO,
  title={GenEval: An Object-Focused Framework for Evaluating Text-to-Image Alignment},
  author={Dhruba Ghosh and Hanna Hajishirzi and Ludwig Schmidt},
  journal={ArXiv},
  year={2023},
  volume={abs/2310.11513},
  url={https://api.semanticscholar.org/CorpusID:264288728}
}

@article{Hu2024ELLAED,
  title={ELLA: Equip Diffusion Models with LLM for Enhanced Semantic Alignment},
  author={Xiwei Hu and Rui Wang and Yixiao Fang and Bin Fu and Pei Cheng and Gang Yu},
  journal={ArXiv},
  year={2024},
  volume={abs/2403.05135},
  url={https://api.semanticscholar.org/CorpusID:268296755}
}

@article{Niu2025WISEAW,
  title={WISE: A World Knowledge-Informed Semantic Evaluation for Text-to-Image Generation},
  author={Yuwei Niu and Munan Ning and Mengren Zheng and Bin Lin and Peng Jin and Jiaqi Liao and Kunpeng Ning and Bin Zhu and Li Yuan},
  journal={ArXiv},
  year={2025},
  volume={abs/2503.07265},
  url={https://api.semanticscholar.org/CorpusID:276929205}
}

@article{Zhao2025EnvisioningBT,
  title={Envisioning Beyond the Pixels: Benchmarking Reasoning-Informed Visual Editing},
  author={Xiangyu Zhao and Peiyuan Zhang and Kexian Tang and Hao Li and Zicheng Zhang and Guangtao Zhai and Junchi Yan and Hua Yang and Xue Yang and Haodong Duan},
  journal={ArXiv},
  year={2025},
  volume={abs/2504.02826},
  url={https://api.semanticscholar.org/CorpusID:277510499}
}

@article{Meng2024PhyBenchAP,
  title={PhyBench: A Physical Commonsense Benchmark for Evaluating Text-to-Image Models},
  author={Fanqing Meng and Wenqi Shao and Li Ray Luo and Yahong Wang and Yiran Chen and Quanfeng Lu and Yue Yang and Tianshuo Yang and Kaipeng Zhang and Yu Qiao and Ping Luo},
  journal={ArXiv},
  year={2024},
  volume={abs/2406.11802},
  url={https://api.semanticscholar.org/CorpusID:270560653}
}

@misc{openai2024gptimage1,
  author       = {OpenAI},
  title        = {GPT-Image-1: Models and capabilities for image generation},
  year         = {2024},
  howpublished = {\url{https://platform.openai.com/docs/models/gpt-image-1}},
  note         = {Accessed: 2026-01-29}
}

@misc{openai2025gptimage15,
  author       = {OpenAI},
  title        = {GPT-Image-1.5: Enhanced visual reasoning and creative generation},
  year         = {2025},
  howpublished = {\url{https://platform.openai.com/docs/models/gpt-image-1.5}},
  note         = {Accessed: 2026-01-29}
}

@misc{blackforestlabs2026flux2pro,
  title        = {FLUX 2 pro: State-of-the-art quality at maximum speed.},
  author       = {{Black Forest Labs}},
  year         = {2026},
  howpublished = {\url{https://bfl.ai/models/flux-2}},
}

@misc{blackforestlabs2026flux2max,
  title        = {FLUX 2 Max: Next Generation Image Synthesis},
  author       = {{Black Forest Labs}},
  year         = {2026},
  howpublished = {\url{https://bfl.ai/models/flux-2-max}},
}

@misc{deepmind2024geminiimage,
  title        = {Gemini Image Pro: High-quality image generation},
  author       = {{Google DeepMind}},
  year         = {2025},
  howpublished = {\url{https://deepmind.google/models/gemini-image/pro/}},
  note         = {Accessed: 2026-01-26}
}

@misc{deepmind2024geminiimage25,
  title        = {Gemini Image: High-quality image generation},
  author       = {{Google DeepMind}},
  year         = {2025},
  howpublished = {\url{https://deepmind.google/models/gemini-image/flash/}},
  note         = {Accessed: 2026-01-26}
}

@article{podell2023sdxl,
  title={Sdxl: Improving latent diffusion models for high-resolution image synthesis},
  author={Podell, Dustin and English, Zion and Lacey, Kyle and Blattmann, Andreas and Dockhorn, Tim and M{\"u}ller, Jonas and Penna, Joe and Rombach, Robin},
  journal={arXiv preprint arXiv:2307.01952},
  year={2023}
}

@misc{sd35Medium,
  title        = {Stable Diffusion 3.5 Medium},
  author       = {{Stability AI}},
  year         = {2024},
  howpublished = {\url{https://huggingface.co/stabilityai/stable-diffusion-3.5-medium/}},
}

@misc{sd35Large,
  title        = {Stable Diffusion 3.5 Large},
  author       = {{Stability AI}},
  year         = {2024},
  howpublished = {\url{https://huggingface.co/stabilityai/stable-diffusion-3.5-large}},
}

@misc{sd3Medium,
  title        = {Stable Diffusion 3 Medium},
  author       = {{Stability AI}},
  year         = {2024},
  howpublished = {\url{https://huggingface.co/stabilityai/stable-diffusion-3-medium}},
}

@misc{flux2024,
  author={Black Forest Labs},
  title={FLUX},
  year={2024},
  howpublished={\url{https://github.com/black-forest-labs/flux}},
}

@misc{labs2025flux1kontextflowmatching,
  title={FLUX.1 Kontext: Flow Matching for In-Context Image Generation and Editing in Latent Space},
  author={Black Forest Labs and Stephen Batifol and Andreas Blattmann and Frederic Boesel and Saksham Consul and Cyril Diagne and Tim Dockhorn and Jack English and Zion English and Patrick Esser and Sumith Kulal and Kyle Lacey and Yam Levi and Cheng Li and Dominik Lorenz and Jonas Müller and Dustin Podell and Robin Rombach and Harry Saini and Axel Sauer and Luke Smith},
  year={2025},
  eprint={2506.15742},
  archivePrefix={arXiv},
  primaryClass={cs.GR},
  url={https://arxiv.org/abs/2506.15742},
}

@article{deng2025bagel,
  title={Emerging properties in unified multimodal pretraining},
  author={Deng, Chaorui and Zhu, Deyao and Li, Kunchang and Gou, Chenhui and Li, Feng and Wang, Zeyu and Zhong, Shu and Yu, Weihao and Nie, Xiaonan and Song, Ziang and others},
  journal={arXiv preprint arXiv:2505.14683},
  year={2025}
}

@article{ye2025echo,
  title={Echo-4o: Harnessing the power of gpt-4o synthetic images for improved image generation},
  author={Ye, Junyan and Jiang, Dongzhi and Wang, Zihao and Zhu, Leqi and Hu, Zhenghao and Huang, Zilong and He, Jun and Yan, Zhiyuan and Yu, Jinghua and Li, Hongsheng and others},
  journal={arXiv preprint arXiv:2508.09987},
  year={2025}
}

@article{jiang2025draco,
  title={DraCo: Draft as CoT for Text-to-Image Preview and Rare Concept Generation},
  author={Jiang, Dongzhi and Zhang, Renrui and Li, Haodong and Zong, Zhuofan and Guo, Ziyu and He, Jun and Guo, Claire and Ye, Junyan and Fang, Rongyao and Li, Weijia and others},
  journal={arXiv preprint arXiv:2512.05112},
  year={2025}
}

@article{cai2025z,
  title={Z-Image: An Efficient Image Generation Foundation Model with Single-Stream Diffusion Transformer},
  author={Cai, Huanqia and Cao, Sihan and Du, Ruoyi and Gao, Peng and Hoi, Steven and Hou, Zhaohui and Huang, Shijie and Jiang, Dengyang and Jin, Xin and Li, Liangchen and others},
  journal={arXiv preprint arXiv:2511.22699},
  year={2025}
}

@article{wu2025qwen,
  title={Qwen-image technical report},
  author={Wu, Chenfei and Li, Jiahao and Zhou, Jingren and Lin, Junyang and Gao, Kaiyuan and Yan, Kun and Yin, Sheng-ming and Bai, Shuai and Xu, Xiao and Chen, Yilei and others},
  journal={arXiv preprint arXiv:2508.02324},
  year={2025}
}

@inproceedings{Zhao2026QwenImage20TR,
  title={Qwen-Image-2.0 Technical Report},
  author={Bin Zhao and Chenfei Wu and De-mei Li and Haoliang Meng and Jiahao Li and Jie Zhang and Jingren Zhou and Junyan Lin and Kaiyuan Gao and Kuang Cao and Kun Yan and Liang Peng and Lihan Jiang and Niantong Li and Ningyuan Tang and Shengming Yin and Tianhe Wu and Xiao Xu and Xiaoyu Chen and Xihua Wang and Yan Shu and Yanran Zhang and Yi Wang and Yilei Chen and Ying Ba and Yixian Xu and Yujia Wu and Yuxiang Chen and Zecheng Tang and Zekai Zhang and Zhendong Wang and Zihao Liu and Zikai Zhou and Anke Yang and Chen Cheng and Chenxu Lv and Dayiheng Liu and Fan Zhou and Han Xiong and Hongzhu Shi and Hu Wei and Hui Zhao and Ivy Liu and Jianwei Zhang and Jiawei Zhang and Kai Chen and Kang He and Le Xue and Lin Qu and Li Tang and Lu-Lu Feng and Min Wu and Minmin Sun and Na Ni and Rui Men and Shuai Bai and Si Zheng and Tao Lan and Tianqi Zhang and Tingkun Wen and Wei Wang and Wei Qiao and Weiyi Lu and Wenmeng Zhou and Xiaodong Deng and Xiaoxiao Xu and Xin Yu Fang and Xiong-hui Chen and Yanan Wang and Yang Fan and Yichang Zhang and Yi-Xuan Xu and Yu Wu and Zhiyuan Ma and Zhi Cai},
  year={2026},
  url={https://api.semanticscholar.org/CorpusID:288256176}
}

@article{chen2025janus,
  title={Janus-Pro: Unified Multimodal Understanding and Generation with Data and Model Scaling},
  author={Chen, Xiaokang and Wu, Zhiyu and Liu, Xingchao and Pan, Zizheng and Liu, Wen and Xie, Zhenda and Yu, Xingkai and Ruan, Chong},
  journal={arXiv preprint arXiv:2501.17811},
  year={2025}
}

@article{wu2024janus,
  title={Janus: Decoupling visual encoding for unified multimodal understanding and generation},
  author={Wu, Chengyue and Chen, Xiaokang and Wu, Zhiyu and Ma, Yiyang and Liu, Xingchao and Pan, Zizheng and Liu, Wen and Xie, Zhenda and Yu, Xingkai and Ruan, Chong and others},
  journal={arXiv preprint arXiv:2410.13848},
  year={2024}
}

@inproceedings{rombach2022high,
  title={High-resolution image synthesis with latent diffusion models},
  author={Rombach, Robin and Blattmann, Andreas and Lorenz, Dominik and Esser, Patrick and Ommer, Bj{\"o}rn},
  booktitle={Proceedings of the IEEE/CVF conference on computer vision and pattern recognition},
  pages={10684--10695},
  year={2022}
}

@article{lin2025uniworld,
  title={UniWorld: High-Resolution Semantic Encoders for Unified Visual Understanding and Generation},
  author={Lin, Bin and Li, Zongjian and Cheng, Xinhua and Niu, Yuwei and Ye, Yang and He, Xianyi and Yuan, Shenghai and Yu, Wangbo and Wang, Shaodong and Ge, Yunyang and others},
  journal={arXiv preprint arXiv:2506.03147},
  year={2025}
}

@article{Wu2025QwenImageTR,
  title={Qwen-Image Technical Report},
  author={Chenfei Wu and Jiahao Li and Jingren Zhou and Junyang Lin and Kaiyuan Gao and Kun Yan and Shengming Yin and Shuai Bai and Xiao Xu and Yilei Chen and Yuxiang Chen and Zecheng Tang and Zekai Zhang and Zhengyi Wang and An Yang and Bowen Yu and Chen Cheng and Dayiheng Liu and De-mei Li and Hang Zhang and Hao Meng and Hu Wei and Ji-Li Ni and Kai Chen and Kuang Cao and Liang Peng and Lin Qu and Min Wu and Peng Wang and Shuting Yu and Tingkun Wen and Wensen Feng and Xiao-Xue Xu and Yi Wang and Yichang Zhang and Yong-An Zhu and Yujian Wu and Yu-Jiao Cai and Ze-Yang Liu},
  journal={ArXiv},
  year={2025},
  volume={abs/2508.02324},
  url={https://api.semanticscholar.org/CorpusID:280422608}
}

@inproceedings{Ren2026SCOPESD,
  title={SCOPE: Structured Decomposition and Conditional Skill Orchestration for Complex Image Generation},
  author={Tianfei Ren and Zhipeng Yan and Yiming Zhao and Zhen Fang and Yu Zeng and Guohui Zhang and Hang Xu and Xiaoxiao Ma and Shiting Huang and Ke Xu and Wenxuan Huang and Lionel Z. Wang and Lin Chen and Zehui Chen and Jie Huang and Feng Zhao},
  year={2026},
  url={https://api.semanticscholar.org/CorpusID:288148127}
}

@misc{bytedance_seedream5_lite_2025,
  author       = {{ByteDance Seed}},
  title        = {Deeper Thinking, More Accurate Generation: Introducing Seedream 5.0 Lite},
  year         = {2025},
  howpublished = {\url{https://seed.bytedance.com/zh/blog/deeper-thinking-more-accurate-generation-introducing-seedream-5-0-lite}},
  note         = {Accessed: 2025-06-19}
}
\bibliographystyle{colm2024_conference}

\newpage
\appendix
\section{Appendix}
\subsection{Case Study}
In this section, we present several case studies to demonstrate the agentic image generation capabilities of Qwen-Image-Agent, including Plan, Reason, Search, Memory, and Feedback, supporting both multi-image and multi-turn generation.

\begin{figure*}[h]
    \centering
    \includegraphics[width=\linewidth]{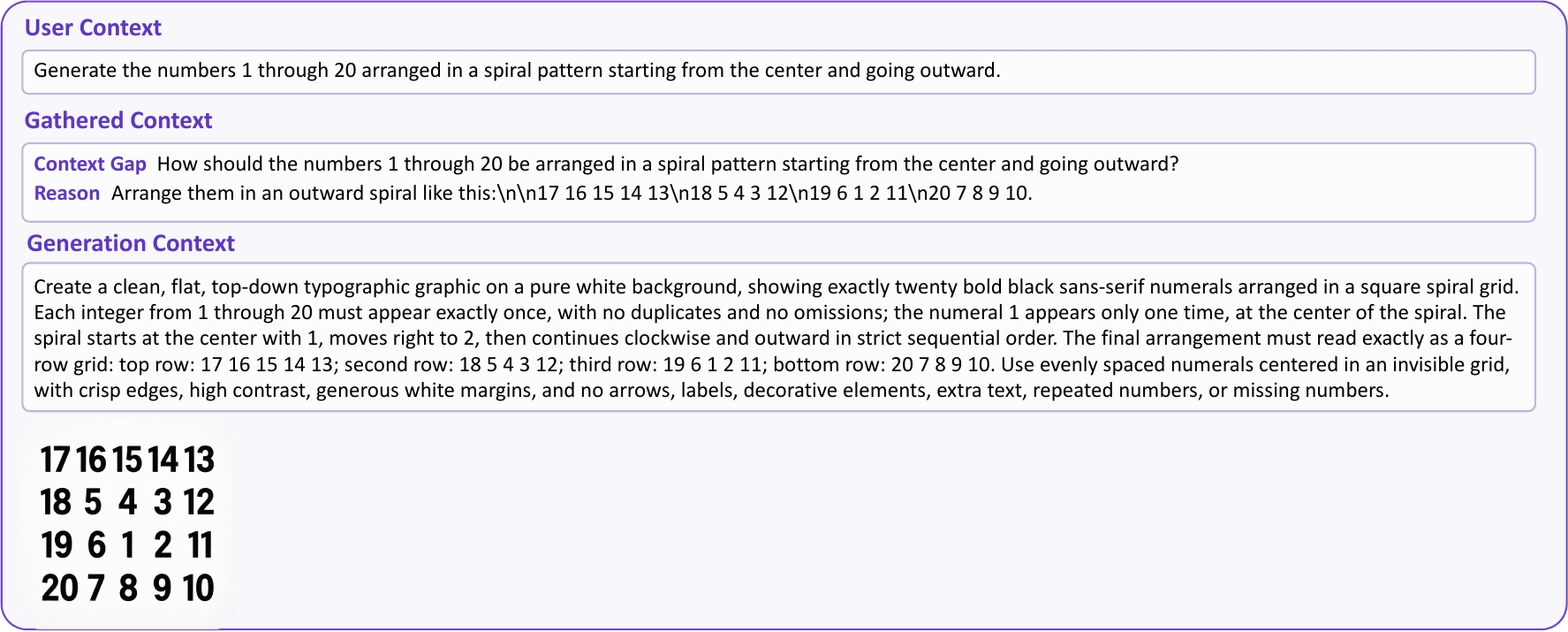}
    \caption{Case Study of planning ability. Qwen-Image-Agent solves the enumeration problem by planning the arrangement.}
    \label{fig:case_plan}
\end{figure*}

\begin{figure*}[h]
    \centering
    \includegraphics[width=\linewidth]{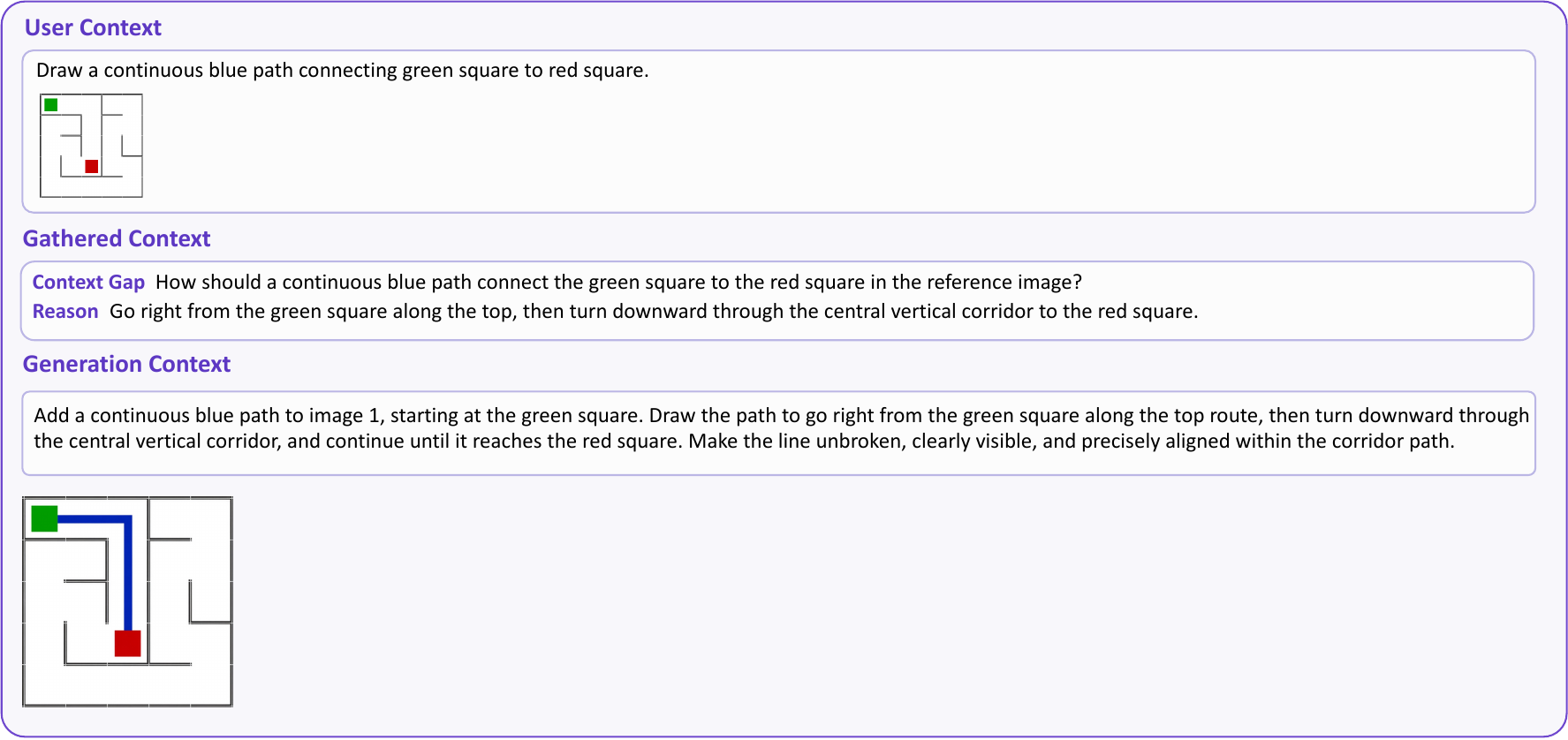}
    \caption{Case Study of reasoning ability. Qwen-Image-Agent solves the maze problem by reasoning the concrete path.}
    \label{fig:case_reason}
\end{figure*}

\begin{figure*}[h]
    \centering
    \includegraphics[width=\linewidth]{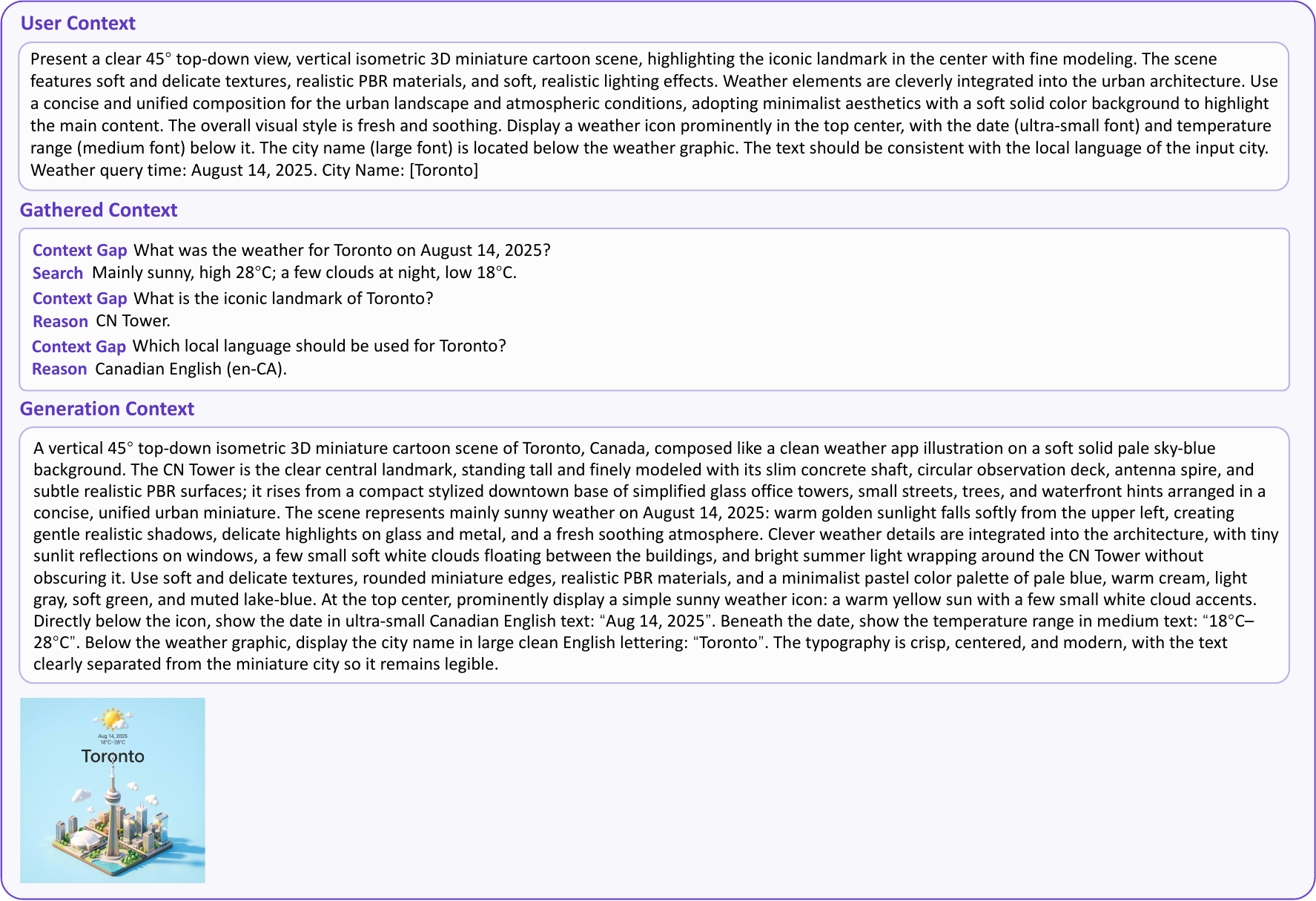}
    \caption{Case Study of web search ability. Qwen-Image-Agent solves the problem by retrieving external knowledge from web.}
    \label{fig:case_web_search}
\end{figure*}

\begin{figure*}[t]
    \centering
    \includegraphics[width=\linewidth]{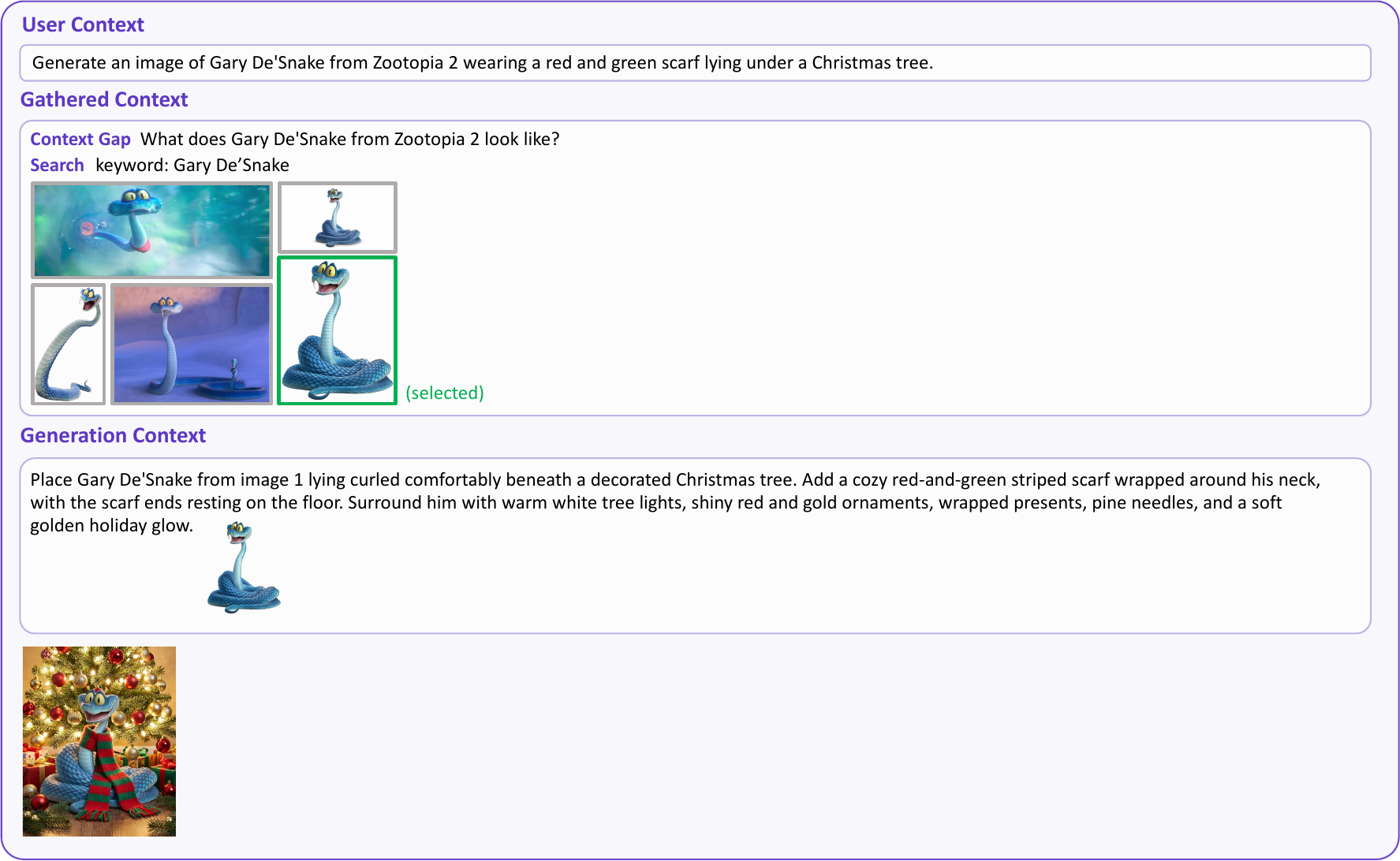}
    \caption{Case Study of image search ability. Qwen-Image-Agent solves the problem by retrieving visual reference from web.}
    \label{fig:case_image_search}
\end{figure*}

\begin{figure*}[t]
    \centering
    \includegraphics[width=\linewidth]{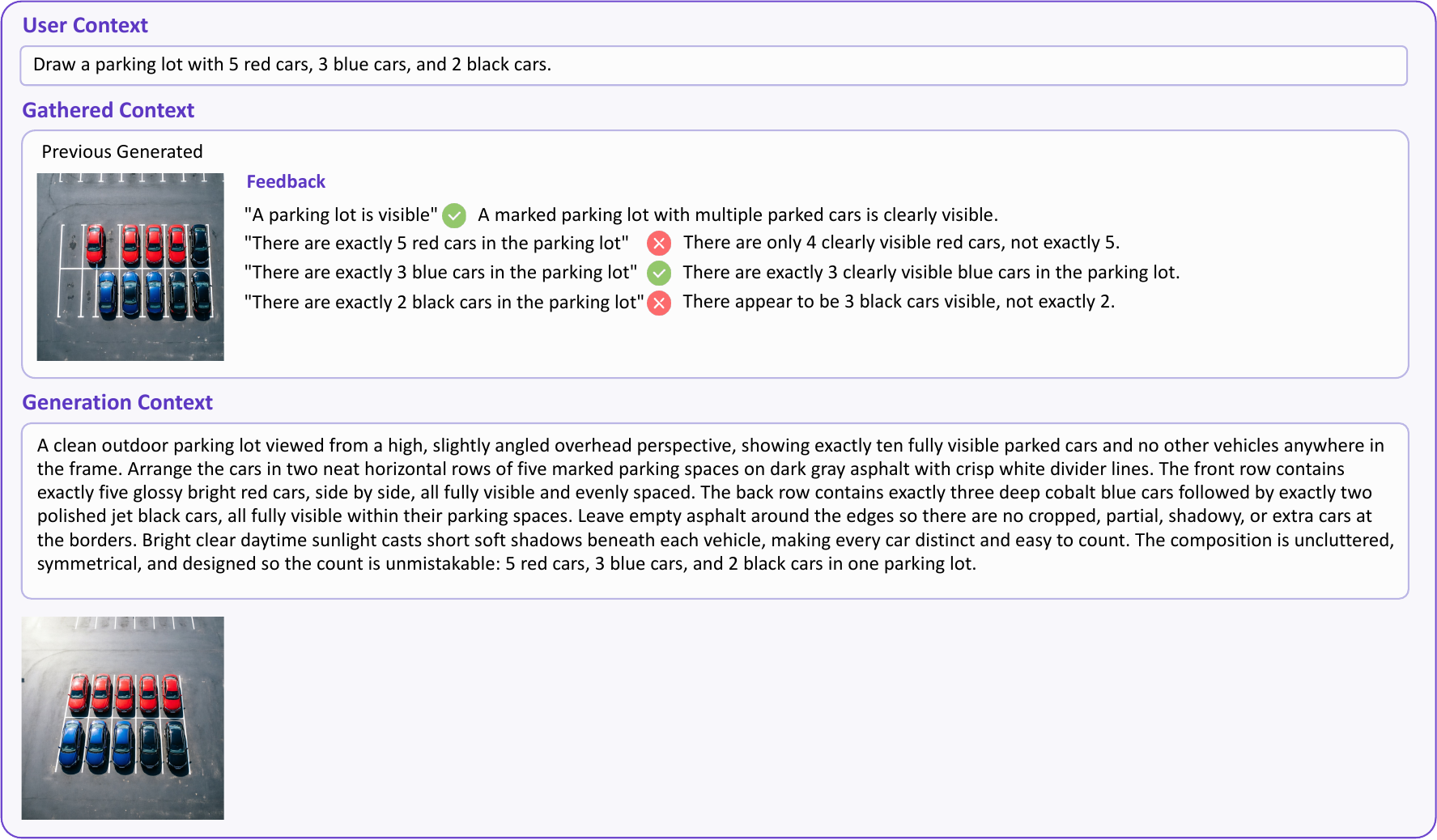}
    \caption{Case Study of feedback ability. Qwen-Image-Agent solves counted composition through self correction.}
    \label{fig:case_feedback}
\end{figure*}

\begin{figure*}[t]
    \centering
    \includegraphics[width=\linewidth]{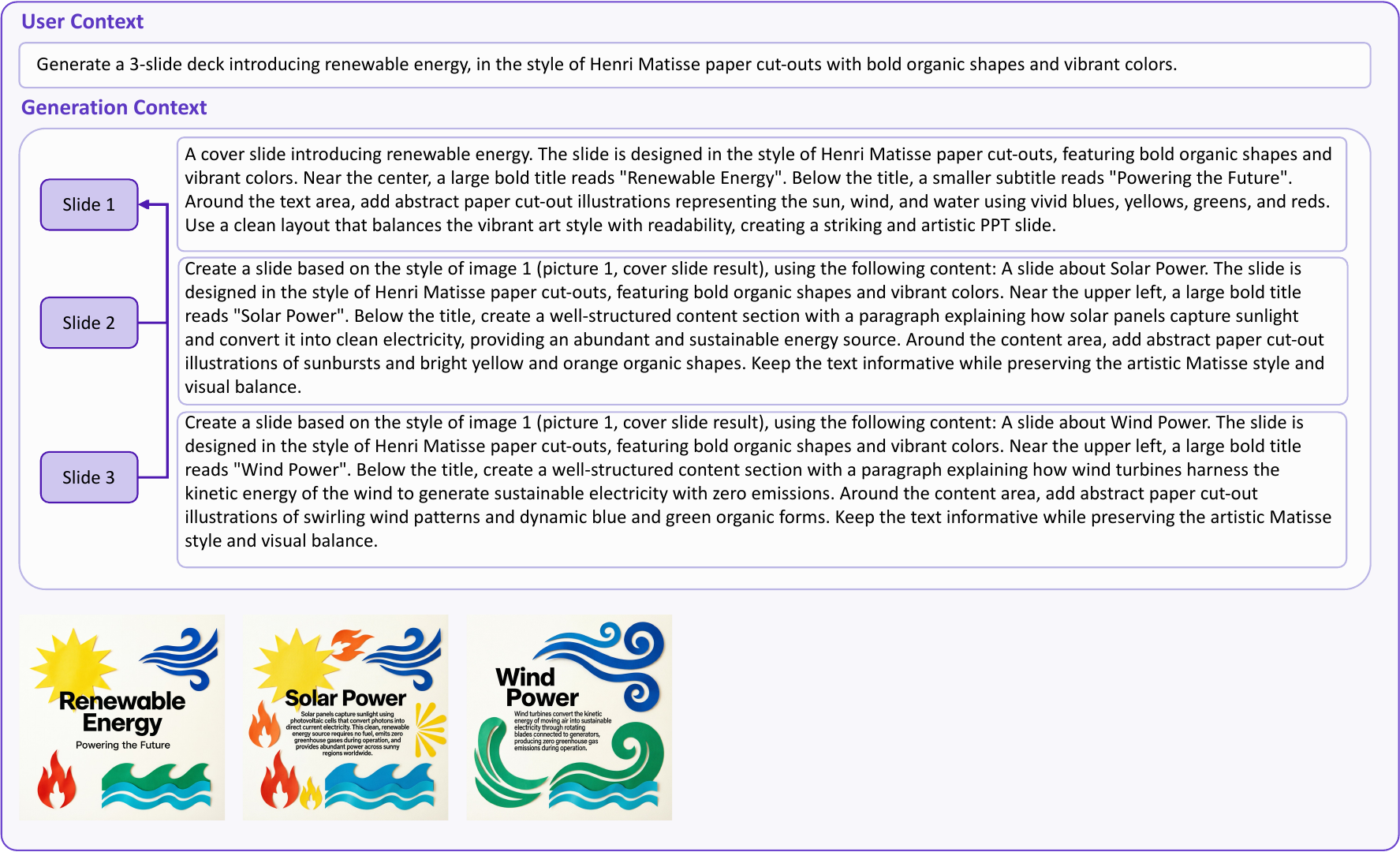}
    \caption{Case Study of multi-image ability. Qwen-Image-Agent enables multi-image generation through splitting and allocating generation context.}
    \label{fig:case_feedback}
\end{figure*}

\begin{figure*}[t]
    \centering
    \includegraphics[width=\linewidth]{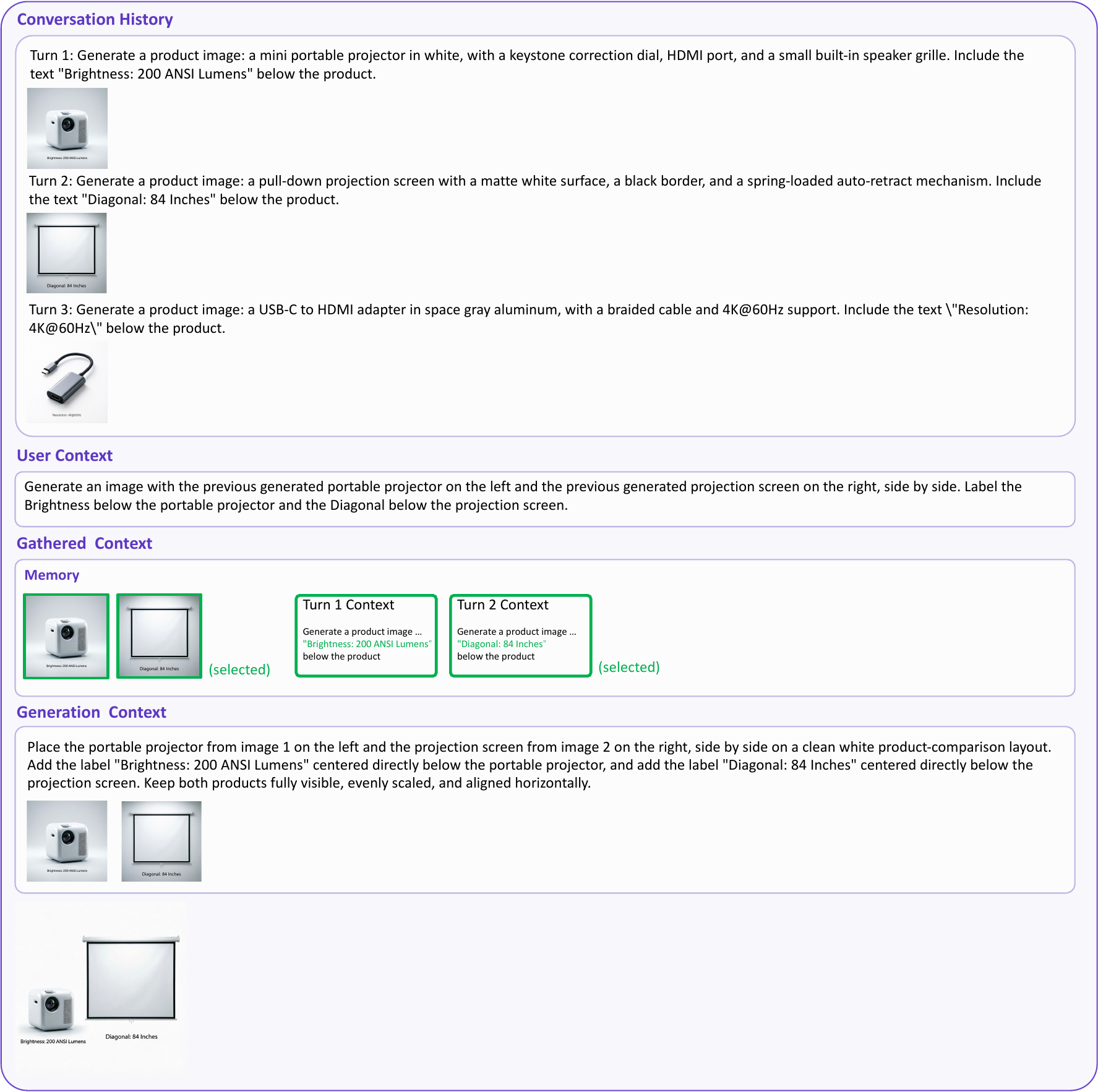}
    \caption{Case Study of memory ability. Qwen-Image-Agent solves the multiturn problem by selecting relevant memory context.}
    \label{fig:case_multiturn}
\end{figure*}

\end{document}